\documentclass[11pt,a4paper]{article}
\usepackage[hyperref]{naaclhlt2019}
\usepackage{times}
\usepackage{latexsym}

\usepackage{url}

\usepackage{amsmath}
\usepackage{amsfonts}
\usepackage{amssymb}
\usepackage{wrapfig}
\usepackage{subcaption}
\usepackage{multirow}
 \usepackage{mathtools} 

\usepackage{verbatim}

\usepackage{anyfontsize}

\usepackage{color}
\usepackage{tikz}
\usetikzlibrary{arrows,shapes,snakes,automata,backgrounds,fit,petri}
\usepackage{adjustbox}

\newcommand{\RN}[1]{%
	\textup{\lowercase\expandafter{\it \romannumeral#1}}%
}

\makeatletter
\newcommand{\distas}[1]{\mathbin{\overset{#1}{\kern\z@\sim}}}%

\usepackage{enumitem}

\usepackage[lined,boxed,commentsnumbered,ruled,linesnumbered]{algorithm2e}

\newcommand{\ie}[0]{\emph{i.e., }}

\newcommand{\eg}[0]{\emph{e.g., }}

\newcommand{\beq}{\vspace{0mm}\begin{equation}}
\newcommand{\eeq}{\vspace{0mm}\end{equation}}
\newcommand{\beqs}{\vspace{0mm}\begin{eqnarray}}
\newcommand{\eeqs}{\vspace{0mm}\end{eqnarray}}
\newcommand{\barr}{\begin{array}}
\newcommand{\earr}{\end{array}}

\newcommand{\cv}[0]{{\boldsymbol{c}}}

\newcommand{\xv}{\boldsymbol{x}}
\newcommand{\yv}{\boldsymbol{y}}
\newcommand{\zv}{\boldsymbol{z}}

\newcommand{\thetav}{\boldsymbol{\theta}}

\newcommand{\phiv}{\boldsymbol{\phi}}
\newcommand{\psiv}{\boldsymbol{\psi}}

\newcommand{\E}{\mathbb{E}}

\newcommand{\Lcal}{\mathcal{L}}

\newcommand{\Fcal}{\mathcal{F}}

\newtheorem{lemma}{Lemma}

\newtheorem{corollary}{Corollary}

\ifx\assumption\undefined
\newtheorem{assumption}{Assumption}
\fi

\ifx\definition\undefined
\newtheorem{definition}{Definition}
\fi

\ifx\remark\undefined
\newtheorem{remark}{Remark}
\fi

\aclfinalcopy % Uncomment this line for the final submission
\def\aclpaperid{1478} %  Enter the acl Paper ID here

\title{Cyclical Annealing Schedule:\\ A Simple Approach to Mitigating KL Vanishing}

\author{Hao Fu$^{1 \dagger} $, ~Chunyuan Li$^{2 \dagger} $\thanks{~ Corresponding author~~ $^{ \dagger}$Equal Contribution}, ~~Xiaodong Liu$^{2}$\\ \bf{Jianfeng Gao}$^{2}$, Asli Celikyilmaz$^{2}$, Lawrence Carin$^{1}$ \\ $^{1}$Duke University ~~~ $^{2}$Microsoft Research, Redmond \\
}

\date{}

\begin{document}
\maketitle
\begin{abstract}
Variational autoencoders (VAEs) with an auto-regressive decoder have been applied for many natural language processing (NLP) tasks. The VAE objective consists of two terms, ($i$) reconstruction and ($ii$) KL regularization, balanced by a weighting hyper-parameter $\beta$. One notorious training difficulty is that the KL term tends to vanish.  In this paper we study scheduling schemes for $\beta$, and show that KL vanishing is caused by the lack of good latent codes in training the decoder at the beginning of optimization.
To remedy this, we propose a cyclical annealing schedule, which repeats the process of increasing $\beta$ multiple times. This new procedure allows the progressive learning of more meaningful latent codes, by leveraging the informative representations of previous cycles as warm re-starts. The effectiveness of cyclical annealing is validated on a broad range of NLP tasks, including language modeling, dialog response generation and unsupervised language pre-training.
\end{abstract}

\section{Introduction}
Variational autoencoders (VAEs)~\cite{kingma2013auto,rezende2014stochastic} have been applied in many NLP tasks, including language modeling~\cite{bowman2015generating,miao2016neural}, dialog response generation~\cite{zhao2017learning,wen2017latent}, semi-supervised text classification~\cite{xu2017variational}, controllable text generation~\cite{hu2017toward}, and text compression~\cite{miao2016language}.
A prominent component of a VAE is the distribution-based latent representation for text sequence observations. 
This flexible representation allows the VAE to explicitly model holistic properties of sentences, such as style, topic, and high-level linguistic and semantic features. 
Samples from the prior latent distribution can produce diverse and well-formed sentences through simple deterministic decoding~\cite{bowman2015generating}.

Due to the sequential nature of text, an auto-regressive decoder is typically employed in the VAE. This is often implemented with a recurrent neural network (RNN); the long short-term memory (LSTM)~\cite{hochreiter1997long} RNN is used widely. This introduces one notorious issue when a VAE is trained using traditional methods: the decoder ignores the latent variable, yielding what is termed the {\it KL vanishing} problem. 

Several attempts have been made to ameliorate this issue~\cite{yang2017improved,dieng2018avoiding,zhao2017learning,kim2018semi}.
Among them, perhaps the simplest solution is monotonic KL annealing, where the weight of the KL penalty term is scheduled to gradually increase during training~\cite{bowman2015generating}.
While these techniques can effectively alleviate the KL-vanishing issue, a proper unified theoretical interpretation is still lacking, even for the simple annealing scheme.

In this paper, we analyze the variable dependency in a VAE, and point out that the auto-regressive decoder has two paths (formally defined in Section~\ref{sec:kl_source}) that work together to generate text sequences. One path is conditioned on the latent codes, and the other path is conditioned on previously generated words. 
% KL vanishing happens because the quality of the latent codes is low at the beginning of the training, thus forcing the VAE to go after the second path, which is easier to optimize. As a result, the latent codes do not encode any useful information for generating the text.
%
KL vanishing happens because 
$(\RN{1})$ the first path can easily get blocked, due to the lack of good latent codes at the beginning of decoder training; 
$(\RN{2})$ the easiest solution that an expressive decoder can learn is to ignore the latent code, and relies on the other path only for decoding.
%
% the easiest solution that 
%
%
To remedy this issue, a promising approach is to 
remove the blockage in the first path, and feed meaningful latent codes in training the decoder, so that the decoder can easily adopt them to generate controllable observations~\cite{bowman2015generating}. 

% one promising approach is to force the VAE to utilize the latent codes for text reconstruction so as to pack useful information into the codes.

This paper makes the following contributions: 
$(\RN{1})$  We provide a novel explanation for the KL-vanishing issue, and develop an understanding of the strengths and weaknesses of existing scheduling methods ($e.g.$, constant or monotonic annealing schedules).
$(\RN{2})$  Based on our explanation, we propose a cyclical annealing schedule. It repeats the annealing process multiple times, and can be considered as an inexpensive approach to leveraging good latent codes learned in the previous cycle, as a warm restart, to train the decoder in the next cycle. 
%
% $(\RN{3})$  We argue that the typically adopted KL term is not informative enough to measure the quality of learned latent codes, and propose the refined metrics for evaluations.
%
$(\RN{3})$ We demonstrate that the proposed cyclical annealing schedule for VAE training improves performance on a large range of tasks (with negligible extra computational cost), including text modeling, dialog response generation, and unsupervised language pre-training.

\section{Preliminaries}
\subsection{The VAE model}
To generate a text sequence of length $T$, $\xv =[x_1, \cdots, x_T]$, neural language models~\cite{mikolov2010recurrent} generate every token $x_t$ conditioned on the previously generated tokens:
\begin{align}
\label{eq_lm_generator}
p(\xv) = \prod_{t=1}^{T} p(x_t| x_{<t}), \nonumber 
\end{align}
where $x_{<t}$ indicates all tokens before $t$.

The VAE model for text consists of two parts, generation and inference~\cite{kingma2013auto,rezende2014stochastic,bowman2015generating}. The {\it generative model} ({\it decoder}) draws a continuous latent vector $\zv$ from prior
$p(\zv)$, and generates the text sequence $\xv$ from a conditional distribution $p_{\thetav}(\xv|\zv)$; $p(\zv)$ is typically assumed a multivariate Gaussian, and $\thetav$ represents the neural network parameters. The following auto-regressive decoding process is usually used: 
\begin{align}
p_{\thetav}(\xv |\zv) = \prod_{t=1}^{T} p_{\thetav}(x_t| x_{<t}, \zv).
% \label{eq_vae_decoder}
\end{align}
Parameters $\thetav$ are typically learned by maximizing the marginal log likelihood $\log p_{\thetav}(\xv) =
\log \int p(\zv) p_{\thetav}(\xv |\zv) \mbox{d} \zv$.   
However, this marginal term is intractable to compute for many decoder choices. Thus, variational inference is considered, and the true posterior $p_{\thetav} (\zv | \xv) \propto p_{\thetav} (\xv | \zv) p(\zv) $ is approximated via
the variational distribution $q_{\phiv}(\zv | \xv)$ is (often known as the {\it inference model} or {\it encoder}), implemented via a $\phiv$-parameterized neural network. 
It yields the {\it evidence lower bound} (ELBO) as an objective:
\begin{align}
&\log p_{\thetav}(\xv) \ge \Lcal_{ \text{ELBO} } = \label{eq_vae_elbo}\\
& 
\E_{q_{\phiv}(\zv | \xv)} \big[ \log p_{\thetav}(\xv | \zv) \big]
-\mbox{KL} (q_{\phiv}(\zv | \xv) || p(\zv) ) \nonumber
\end{align}
Typically, $q_{\phiv} (\zv | \xv)$ is modeled as a Gaussian distribution, and the re-parametrization trick is used for efficient learning~\cite{kingma2013auto}.

\subsection{Training Schedules and KL Vanishing}
There is an alternative interpretation of the ELBO: the VAE objective can be viewed as a regularized version of the autoencoder (AE)~\cite{goodfellow2016deep}. 
It is thus natural to extend the negative of $ \Lcal_{ \text{ELBO} }$ in \eqref{eq_vae_elbo} by introducing a hyper-parameter $\beta$ to control the strength of regularization:
\begin{align}
\Lcal_{\beta}  
& = \Lcal_{E} +\beta \Lcal_{R},~~\text{with} \\
\Lcal_{E} & = -\E_{q_{\phiv}(\zv | \xv)} \big[ \log p_{\thetav}(\xv | \zv) \big]  \\
\Lcal_{R} & = \mbox{KL} (q_{\phiv}(\zv | \xv) || p(\zv) ) 
\label{eq_reg_elbo} 
\end{align}
where $\Lcal_{E}$ is the reconstruction error (or negative log-likelihood (NLL)), and $\Lcal_{R}$ is a KL regularizer.

The cost function $\Lcal_{\beta}$ provides a unified perspective for understanding various autoencoder variants and training methods.
When $\beta = 1$, we recover the VAE in ~\eqref{eq_vae_elbo}.  
When $\beta = 0$, and $q_{\phiv}(\zv | \xv)$ is a delta distribution, we recover the AE. 
% \JG{please check: when $\beta = 0$, $\Lcal_{R}$ is gone. why does $q_{\phiv}(\zv | \xv)$ matter?}
In other words, the AE does not regularize the variational distribution toward a prior distribution, and there is only a point-estimate to represent the text sequence's latent feature. In practice, it has been found that learning with an AE is prone to overfitting~\cite{bowman2015generating}, or generating plain dialog responses~\cite{zhao2017learning}. Hence, it is desirable to retain meaningful posteriors in real applications.
Two different schedules for $\beta$ have been commonly used for a text VAE.
% Two representative schedules have been employed for $\beta$ in training VAEs.

\paragraph{Constant Schedule} The standard approach is to keep $\beta=1$ fixed during the entire training procedure, as it corresponds to optimizing the true VAE objective. Unfortunately, instability on text analysis has been witnessed, in that the KL
term $\Lcal_{R}$ becomes vanishingly small during training~\cite{bowman2015generating}. This issue causes two undesirable outcomes:
$(\RN{1})$ an encoder that produces posteriors almost identical to the Gaussian prior, for all observations (rather than a more interesting posterior); and 
$(\RN{2})$ a decoder that completely ignores the latent variable $\zv$, and a learned model that reduces to a simpler language model. This is known as the {\it KL vanishing} issue in text VAEs. 

\paragraph{Monotonic Annealing Schedule.} A simple remedy has been proposed in~\cite{bowman2015generating} to alleviate KL collapse. It sets $\beta=0$ at the beginning of training, and gradually increases $\beta$ until $\beta=1$ is reached.
In this setting, we do not optimize the proper lower bound in \eqref{eq_vae_elbo} during the early stages of training, but nonetheless improvements
on the value of that bound are observed at convergence in previous work~\cite{bowman2015generating,zhao2017learning}.

The monotonic annealing schedule has become the {\em de facto} standard in training text VAEs, and has been widely adopted in many NLP tasks. Though simple and often effective, this heuristic still lacks a proper justification. Further, how to best schedule $\beta$ is largely unexplored.

% \JG{Is lack of theoretical justification the ONLY motivation for you to propose the new method? The new method does not have any theoretical justification either.}

\begin{figure}[t!]%\vspace{-25pt}
	\vspace{-0mm}\centering
	\begin{tabular}{c c}
		\hspace{-4mm}
		\includegraphics[height=1.1cm]{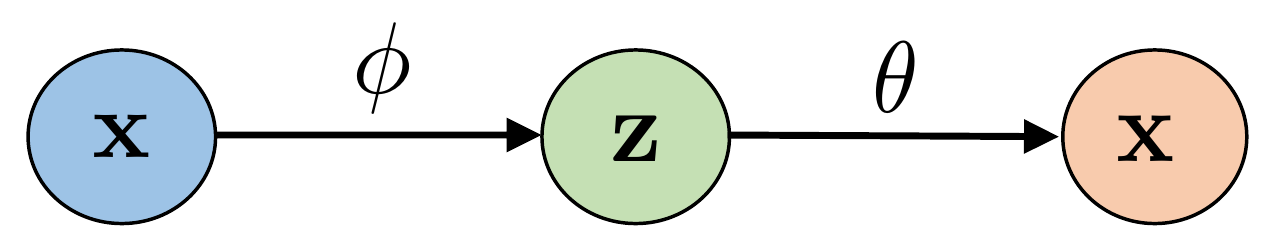} \\
		(a) Traditional VAE \vspace{2mm} \\
		% \vspace{2mm}
		\includegraphics[height=3.2cm]{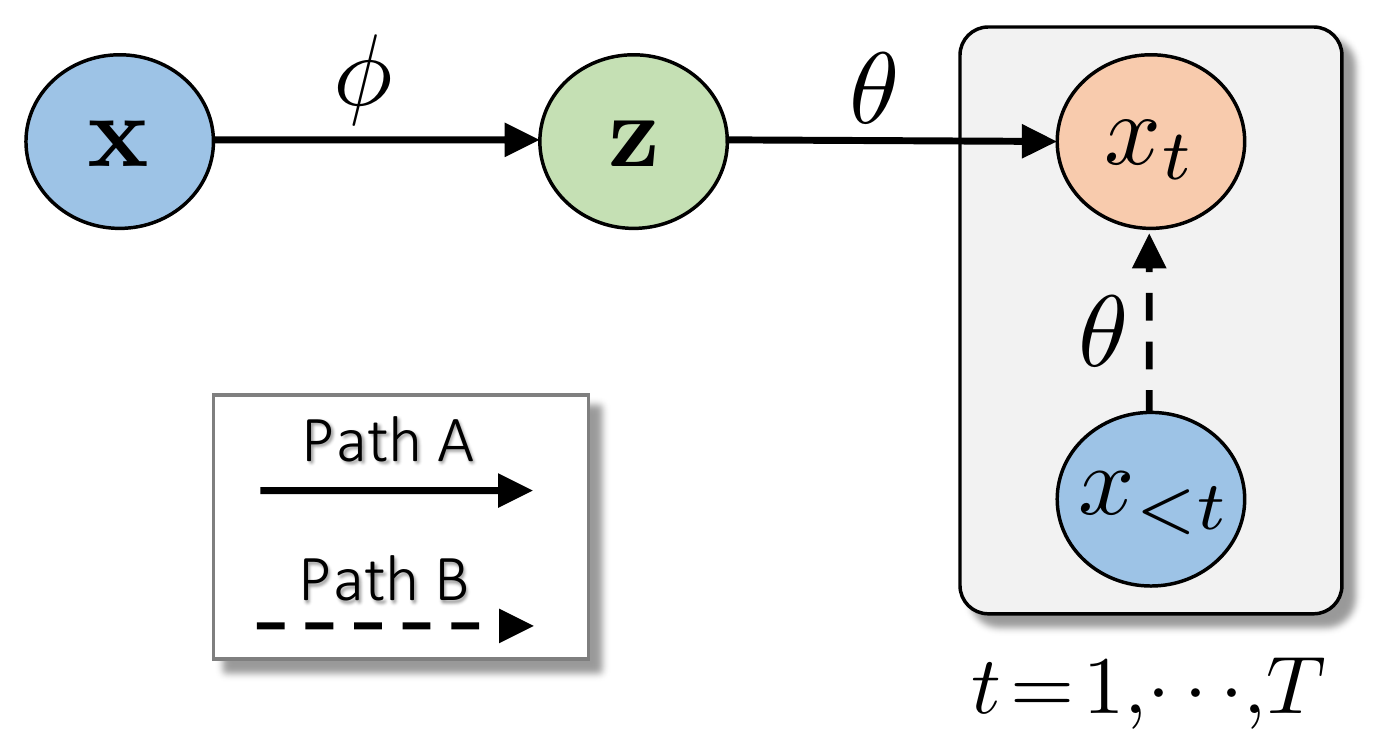} \\
		(b) VAE with an auto-regressive decoder \hspace{-0mm} 
	\end{tabular}
	\vspace{-2mm}
	\caption{Illustration of learning parameters $\{\phiv,\thetav\}$ in the two different paradigms. 
	Starting from the observation $\xv$ in blue circle, a VAE infers its latent code $\zv$ in the green circle, and further generates its reconstruction in the red circle.
	(a) Standard VAE learning, with only one path via $\{\phiv,\thetav\}$ from $\xv$ to its reconstruction; (b) VAE learning with an auto-regressive decoder.  Two paths are considered from $\xv$ to its reconstruction: Path A via $\{\phiv,\thetav\}$ and Path B via $\thetav$. }
	\vspace{-4mm}
	\label{fig:schemes}
\end{figure}

\section{Cyclical Annealing Schedule}
\vspace{-0mm}
\subsection{Identifying Sources of KL Vanishing}
\label{sec:kl_source}

In the traditional VAE~\cite{kingma2013auto}, $\zv$ generates $\xv$ directly, and the reconstruction depends only on one path of $\{\phiv, \thetav\}$ passing through $\zv$, as shown in Figure~\ref{fig:schemes}(a). Hence, $\zv$ can largely determine the reconstructed $\xv$.
%
% The traditional VAE learns to reconstruct the observation $\xv$ through the path of $\{\phiv, \thetav\}$ via $\zv$, and $, as shown in Figure~\ref{fig:schemes}(a). $\zv$ largely determines the reconstructed $\xv$. 
%
In contrast, when an auto-regressive decoder is used in a text VAE~\cite{bowman2015generating}, there are two paths from $\xv$ to its reconstruction, as shown in Figure~\ref{fig:schemes}(b). 
%
% \JG{It is confusing. Figure 1(a) is an autoencoder, and Figure 1(b) is VAE. I don't know what ``traditional VAE'' is. Can you give a reference for ``VAE with an auto-regressive decoder''? Or you need to explain explicitly that in this paper we explore VAE for text generation. Due to the discrete and sequential nature of text, we use RNN or LSTM in VAE -- called VAE with auto-regressive decoder (?)}
%
Path A is the same as that in the standard VAE, where $\zv$ is the global representation that controls the generation of $\xv$; 
Path B leaks the partial ground-truth information of $\xv$ at every time step of the sequential decoding. It generates $x_t$ conditioned on $x_{<t}$. Therefore, Path B can potentially bypass Path A to generate $\xv$, leading to KL vanishing. 

From this perspective, we hypothesize that the model-collapse problem is related to the low quality of $\zv$ 
% in training the decoder 
at the beginning phase of decoder training. A lower quality $\zv$ introduces more difficulties in reconstructing $\xv$ via Path A. As a result, the model is forced to learn an easier solution to decoding: generating $\xv$ via Path B only. 

We argue that this phenomenon can be easily observed due to the powerful representation capability of the auto-regressive decoder. It has been shown empirically that auto-regressive decoders are able to capture highly-complex distributions, such as natural language sentences~\cite{mikolov2010recurrent}. This means that Path B alone has enough capacity to model $\xv$, even though the decoder takes $\{x_{<t}, \zv \}$ as input to produce $x_t$. 
\citet{zhang2017understanding} has shown that flexible deep neural networks can easily fit randomly labeled training data, and here the decoder can learn to rely solely on $ x_{<t} $ for generation, when $\zv$ is of low quality.

%Further, our hypothesis can 
We use our hypothesis to explain the learning behavior of different scheduling schemes for $\beta$ as follows. 

\paragraph{Constant Schedule} 
The two loss terms in~\eqref{eq_vae_elbo} are weighted equally in the constant schedule. 
%equally considers the two losses . 
At the early stage of optimization, $\{\phiv,\thetav\}$ are randomly initialized and the latent codes $\zv$ are of low quality.
The KL term $\Lcal_{R}$ pushes $q_{\phiv}(\zv|\xv)$ close to an uninformative prior $p(\zv)$: the posterior becomes more like an isotropic Gaussian noise, and less representative of their corresponding observations. 
In other words, $\Lcal_{R}$ blocks Path A, and thus $\zv$ remains uninformative during the entire training process: it starts with random initialization and then is regularized towards a random noise. 
Although the reconstruction term $\Lcal_{E}$ can be satisfied via two paths, since $\zv$ is noisy, 
% the easiest solution for the flexible 
the decoder learns to discard Path A (\ie ignores $\zv$), and chooses Path B to generate the sentence word-by-word. 
%to directly pass from $x_{<t}$ to $x_t$.

\paragraph{Monotonic Annealing Schedule}
The monotonic schedule sets $\beta$ close to $0$ in the early stage of training, which effectively removes the blockage $\Lcal_{R}$ on Path A, and the model reduces to a denoising autoencoder~\footnote{The Gaussian sampling remains for $q_{\phiv}(\zv|\xv)$}.
$\Lcal_{E}$ becomes the only objective, which can be reached by both paths.
Though randomly initialized, $\zv$ is learned to capture useful information for reconstruction of $\xv$ during training. 
% then optimized to be representative enough to reconstruct .
%
At the time when the full VAE objective is considered ($\beta=1$), $\zv$ learned earlier can be viewed as the VAE initialization; such latent variables are much more informative than random, and thus are ready for the decoder to use. % learns to use $\zv$.

To mitigate the KL-vanishing issue, it is key to have meaningful latent codes $\zv$ at the beginning of training the decoder, so that $\zv$ can be utilized.
The monotonic schedule under-weights the prior regularization, and the learned $q_{\phiv}(\zv|\xv)$ tends to collapse into a point estimate (\ie the VAE reduces to an AE). This underestimate can result in sub-optimal decoder learning.
A natural question concerns how one can get a better distribution estimate for $\zv$ as initialization, while retaining low computational cost. 

\vspace{-0mm}
\subsection{Cyclical Annealing Schedule}
\vspace{-0mm}
Our proposal is to use $\zv \sim q_{\phiv}(\zv|\xv)$, which has been trained under the full VAE objective, as initialization. 
To learn to progressively improve latent representation $\zv$, we propose a cyclic annealing schedule. We start with $\beta=0$, increase $\beta$ at a fast pace, and then stay at $\beta=1$ for subsequent learning iterations. This encourages the model to converge towards the VAE objective, and infers its first raw full latent distribution. 

Unfortunately, Path A is blocked at $\beta=1$. The optimization is then continued at $\beta=0$ again, which perturbs the VAE objective, dislodges it from the convergence, and reopens Path A. Importantly, the decoder is now trained with the latent code from a full distribution $\zv \sim q_{\phiv}(\zv|\xv)$, and both paths are considered. We repeat this process several times to achieve better convergences.

Formally, $\beta$ has the form:
%
% \begin{align}
% \beta(t) = \min(1,\; \frac{t \bmod T}{K_c}), \quad K_c < T
% \label{eq_beta_monotonic}
% \end{align}
%
% \begin{align}
% \beta(t) = f(\bmod ( t-1, \left \lceil R*T/M \right \rceil) )
% \label{eq_beta_cyclic}
% \end{align}
%
\begin{align}
\beta_t = 
\left\{\begin{matrix}
f(\tau), & \tau \le R\\ 
1,& \tau > R
\end{matrix}\right.
~~~\text{with}~~\\
& \hspace{-25mm}\tau = \frac{  \bmod ( t-1, \left \lceil T/M \right \rceil )  }{ T/M }~,
\label{eq_beta_cyclic}
\end{align}
where $t$ is the iteration number, $T$ is the total number of training iterations, $f$ is a monotonically increasing function, and we introduce two new hyper-parameters associated with the cyclical annealing schedule:
\begin{itemize}
    \item $M$: number of cycles (default $M=4$); 
    \vspace{-2mm}
    \item $R$: proportion used to increase $\beta$ within a cycle  (default $R=0.5$).
\end{itemize}

\begin{figure}[t!]%\vspace{-25pt}
	\vspace{-0mm}\centering
	\begin{tabular}{c}		
		\includegraphics[height=4.5cm]{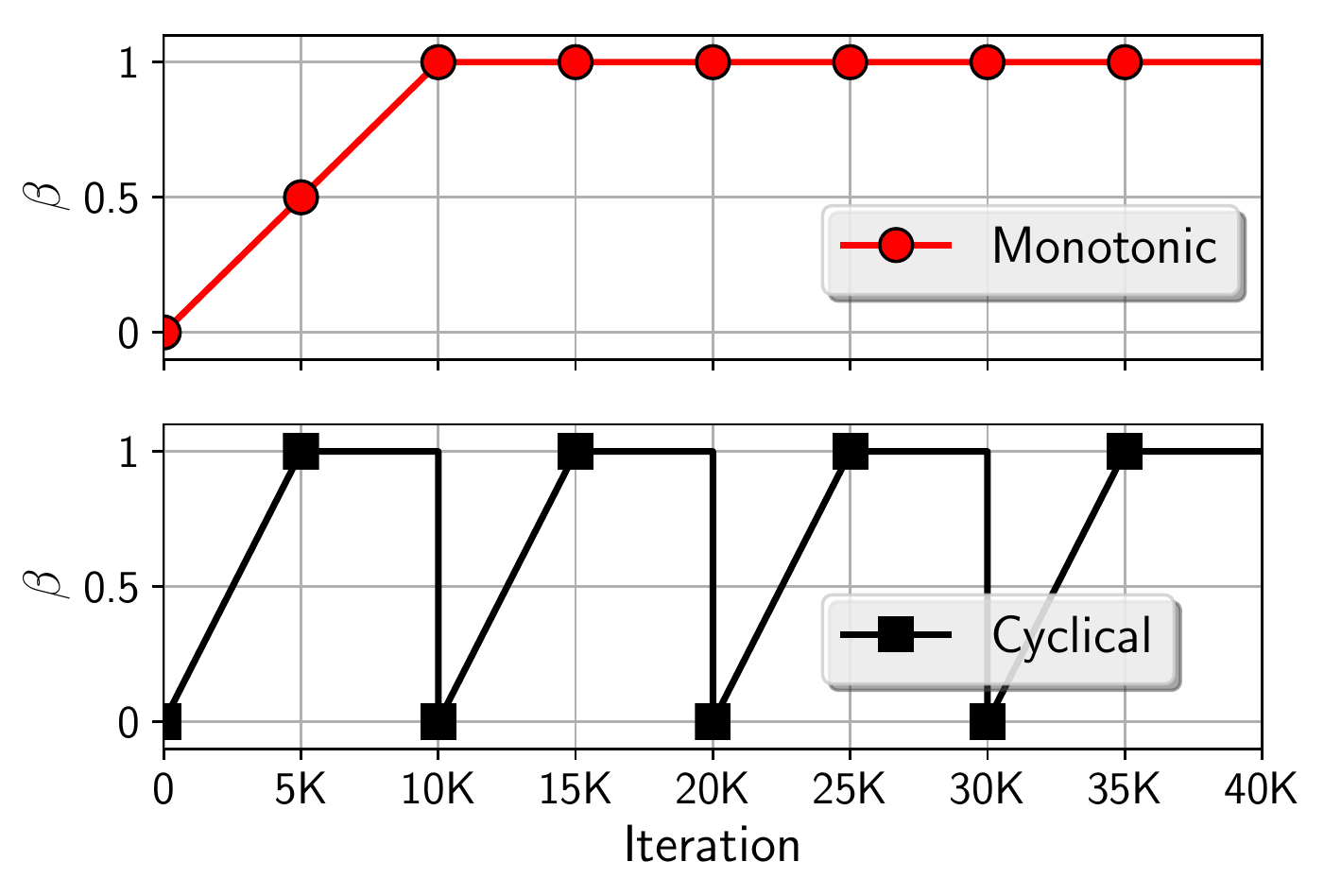} 
	\end{tabular}
	\vspace{-0mm}
	\caption{Comparison between (a) traditional monotonic and (b) proposed cyclical annealing schedules.In this figure, $M=4$ cycles are illustrated, $R=0.5$ is used for increasing within each cycle. }
	\vspace{-0mm}
	\label{fig:schedules}
\end{figure}

In other words, we split the training process into $M$ cycles, each starting with $\beta=0$ and ending with $\beta=1$.
We provide an example of a cyclical schedule in Figure~\ref{fig:schedules}(b), compared with the monotonic schedule in~Figure~\ref{fig:schedules}(a).
Within one cycle, there are two consecutive stages (divided by $R$):
\begin{itemize}
    \item {\bf Annealing}.~ $\beta$ is annealed from $0$ to $1$ in the first $R \left \lceil T/M \right \rceil $ training steps over the course of a cycle. 
    For example, the steps $[1, 5K]$ in the Figure~\ref{fig:schedules}(b).
    $\beta = f(0) = 0$ forces the model to learn representative $\zv$ to reconstruct $\xv$. As depicted in Figure~\ref{fig:schemes}(b), there is no interruption from the prior on Path A, $\zv$ is forced to learn the global representation of $\xv$. By gradually increasing $\beta$ towards $f(R) = 1$, $q(\zv|\xv)$ is regularized to transit from a point estimate to a distribution estimate, spreading out to match the prior.
    \vspace{-2mm}
    \item {\bf Fixing}.~ As our ultimate goal is to learn a VAE model, we fix $\beta=1$ for the rest of training steps within one cycle, \eg the steps $[5K, 10K]$ in Figure~\ref{fig:schedules}(b).
    This drives the model to optimize the full VAE objective until convergence.
\end{itemize}

As illustrated in Figure~\ref{fig:schedules}, the monotonic schedule increasingly anneals $\beta$ from 0 to 1 once, and fixes $\beta=1$ during the rest of training. The cyclical schedules alternatively repeats the annealing and fixing stages multiple times.

\paragraph{A Practical Recipe}~
The existing schedules can be viewed as special cases of the proposed cyclical schedule.
The cyclical schedule reduces to the constant schedule when $R=0$, and it reduces to an monotonic schedule when $M=1$ and $R$ is relatively small~\footnote{In practice, the monotonic schedule usually anneals in a very fast pace, thus $R$ is small compared with the entire training procedure.}.
In theory, any monotonically increasing function $f$ can be adopted for the cyclical schedule, as long as $f(0)=0$ and $f(R)=1$. In practice, we suggest to build the cyclical schedule upon the success of monotonic schedules: we adopt the same $f$, and modify it by setting $M$ and $R$ (as default). Three widely used increasing functions for $f$ are linear~\cite{fraccaro2016sequential,goyal2017z}, Sigmoid~\cite{bowman2015generating} and Consine~\cite{lai2018stochastic}. We present the comparative results using the linear function $f(\tau) = \tau / R $ in Figure~\ref{fig:schedules}, and show the complete comparison for other functions in Figure~\ref{fig:schedules_supp} of the Supplementary Material (SM).

%
% \begin{align}
% \beta(t) = \min(1,\; \frac{t \bmod T}{K_c}), \quad K_c < T
% \label{eq_beta_cycle_linear}
% \end{align}
%

\vspace{-1mm}
\subsection{On the impact of $\beta$}
\vspace{-1mm}
This section derives a bound for the training objective to rigorously study the impact of $\beta$; the proof details are included in SM.
% on reducing KL vanishing. 
% \JG{It is helpful if you sketch how we derive the bound, e.g., by introducing $\beta$ into the ELBO?}
%
For notational convenience, we identify each data sample with a unique integer index $n \sim q(n)$, drawn from a uniform random variable on $\{1,2, \cdots, N \}$. Further we define $q(\zv|n)=q_{\phiv}(\zv|\xv_n)$ and $q(\zv, n) = q(\zv|n)q(n)=q(\zv|n)\frac{1}{N}$.  
Following~\cite{makhzani2016adversarial}, we refer to $q(\zv) = \sum_{n=1}^N q(\zv | n) q(n) $ as the aggregated posterior. This marginal distribution captures the aggregated $\zv$ over the entire dataset.
The KL term in~\eqref{eq_reg_elbo} can be decomposed into two refined terms~\cite{chenisolating,hoffman2016elbo}: 
\vspace{-1mm}
\begin{align} \label{eq_kl_decomp}
 \hspace{-4mm} \Fcal_R & = \E_{q(n)}[ \mbox{KL} (q(\zv | n) || p(\zv) ) ]  \nonumber \\
& =  \underbrace{I_q(\zv, n)}_{\Fcal_1:~\text{Mutual~Info.}} +
\underbrace{ \mbox{KL}(q(\zv) || p(\zv)) }_{\Fcal_2:~\text{Marginal KL} } 
% & \hspace{-4mm} 
% \underbrace{\mbox{KL}(q(\zv) || \prod_j q(z_j))}_{\Fcal_2:~\text{Total Correlation} } + 
% \underbrace{\sum_j \mbox{KL}(q(z_j) || p(z_j))}_{\Fcal_3:~\text{Dimension-wise~KL}}
\end{align}
where 
$\Fcal_1$ is the mutual information (MI) measured by $q$. Higher MI can lead to a higher correlation between the latent variable and data variable, and encourages a reduction in the degree of KL vanishing. The marginal KL is represented by $\Fcal_2$, and it measures the fitness of the aggregated posterior to the prior distribution.

The reconstruction term in~\eqref{eq_reg_elbo} provides a lower bound for MI measured by $q$, based on Corollary 3 in~\cite{li2017alice}:
\vspace{-2mm}
\begin{align} \label{eq_rec_bound}
 \hspace{-2mm} \Fcal_E 
 & = E_{q(n), \zv \sim q(\zv|n)} (\log p(n | \zv ) ) ] + H_q(n) \nonumber \\
 & \le I_q(\zv, n)
\end{align}
where $H(n)$ is a constant. 

\paragraph{Analysis of $\beta$}
When scheduled with $\beta$, the training objective over the dataset can be written as:
\begin{align} \label{eq_ml_analysis}
\Fcal & = -\Fcal_E + \beta \Fcal_R  \\
& \ge (\beta-1) I_q(\zv, n) +  \beta \mbox{KL}(q(\zv) || p(\zv))
\end{align}

To reduce KL vanishing, we desire an increase in the MI term $I(\zv, n)$, which appears in both $\Fcal_E$ and $\Fcal_R$, modulated by $\beta$. It shows that reducing KL vanishing is inversely proportional with $\beta$.
When $\beta=0$, the model fully focuses on maximizing the MI. As $\beta$ increases, the model gradually transits towards fitting the aggregated latent codes to the given prior. When $\beta=1$, the implementation of MI becomes implicit in $\mbox{KL}(q(\zv) || p(\zv))$. It is determined by the amortized inference regularization (implied by the encoder's expressivity)~\cite{shu2018amortized}, which further affects the performance of the generative density estimator.

\begin{figure*}[t!]%\vspace{-25pt}
	\vspace{-0mm}\centering
	\begin{tabular}{ccc}
	    \hspace{-5mm}
		\includegraphics[height=2.7cm]{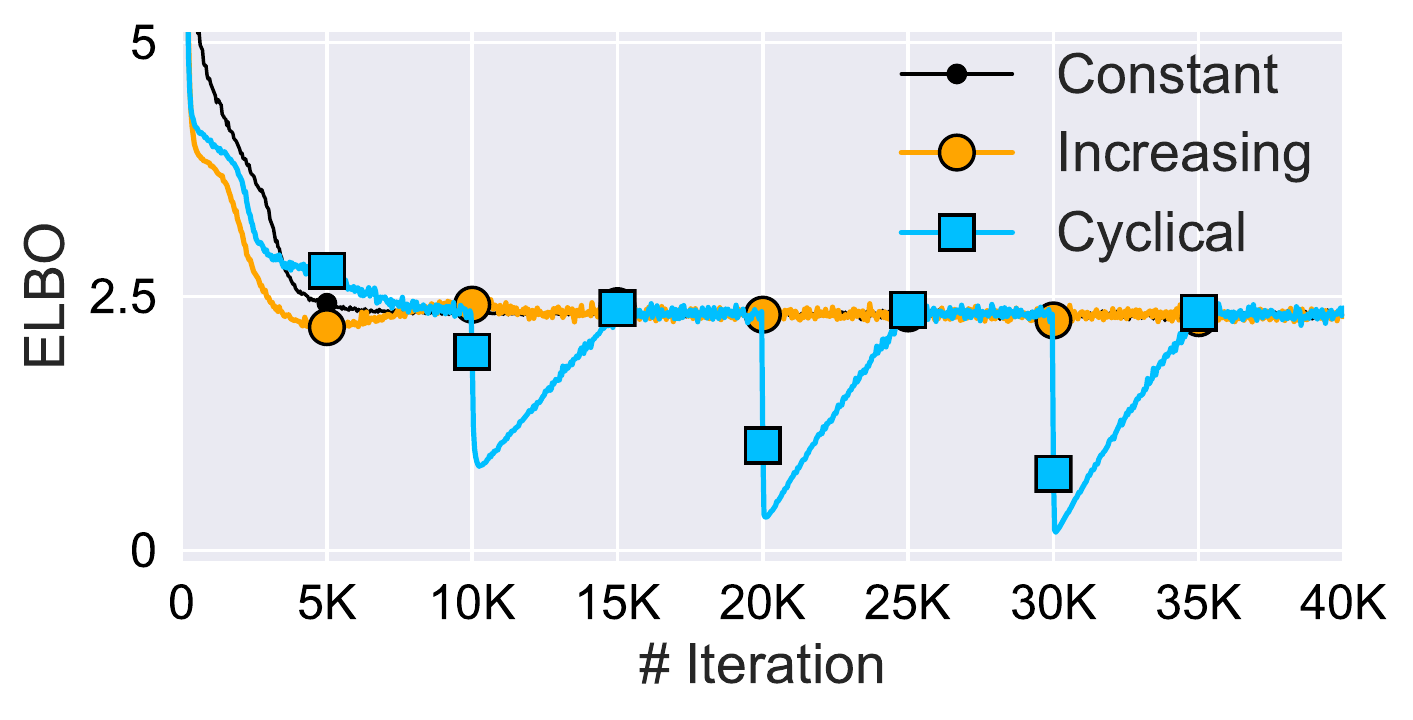} & 
		\hspace{-6mm}
		\includegraphics[height=2.7cm]{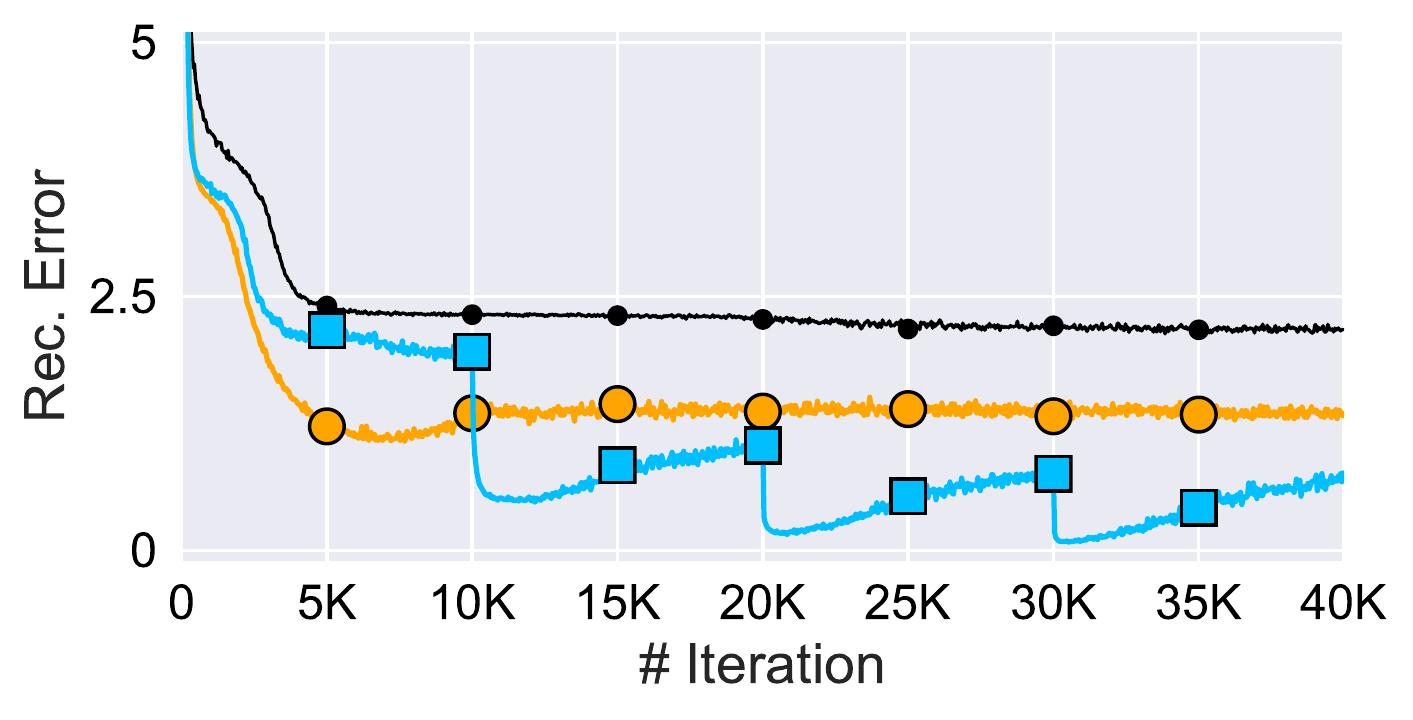} &
		\hspace{-6mm}
		\includegraphics[height=2.7cm]{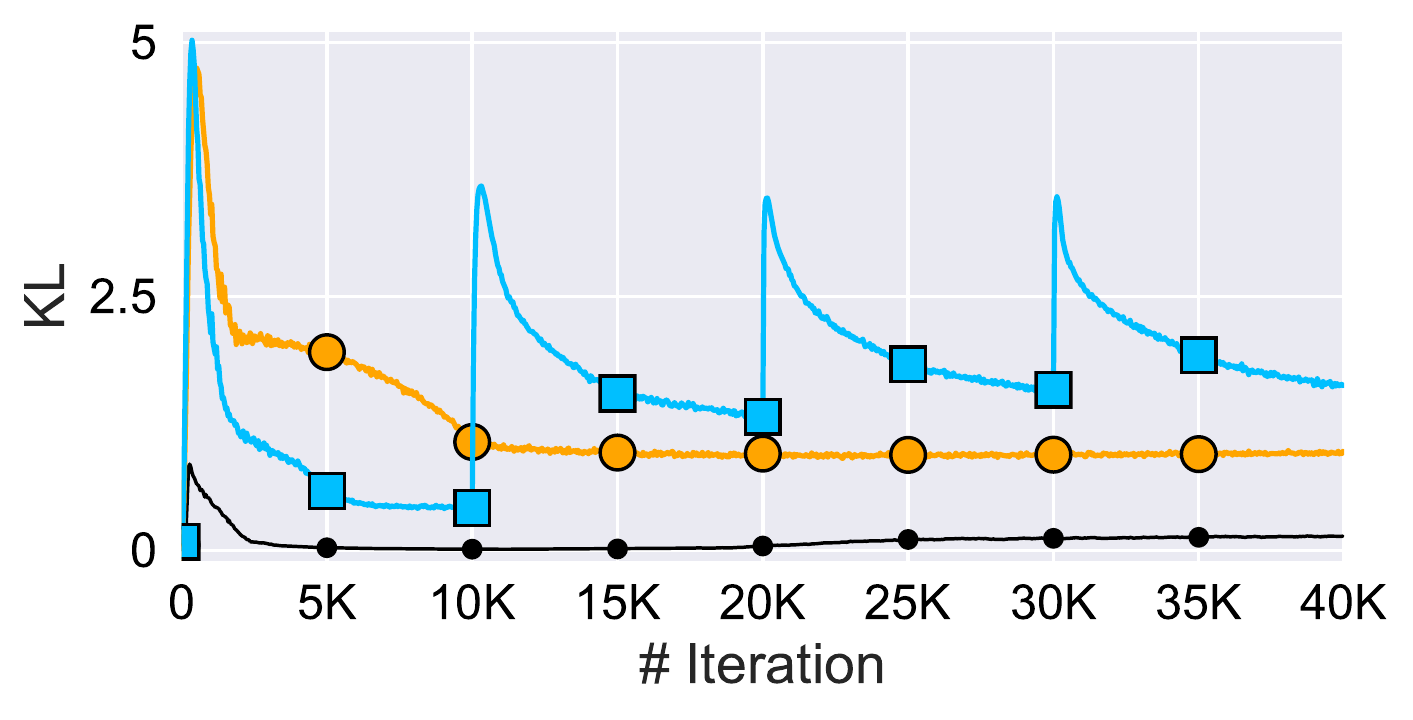} \vspace{-2mm} \\
		(a) ELBO &
		(b) Reconstruction Error &
		(c) KL term
	\end{tabular}
	\vspace{-2mm}
	\caption{Comparison of the learning curves for the three schedules on an illustrative problem.}
	\vspace{-0mm}
	\label{fig:schedules_toy_lc}
\end{figure*}

\begin{figure*}[t!]%\vspace{-25pt}
	\vspace{-0mm}\centering
		\hspace{0mm}
		\includegraphics[height=6.2cm]{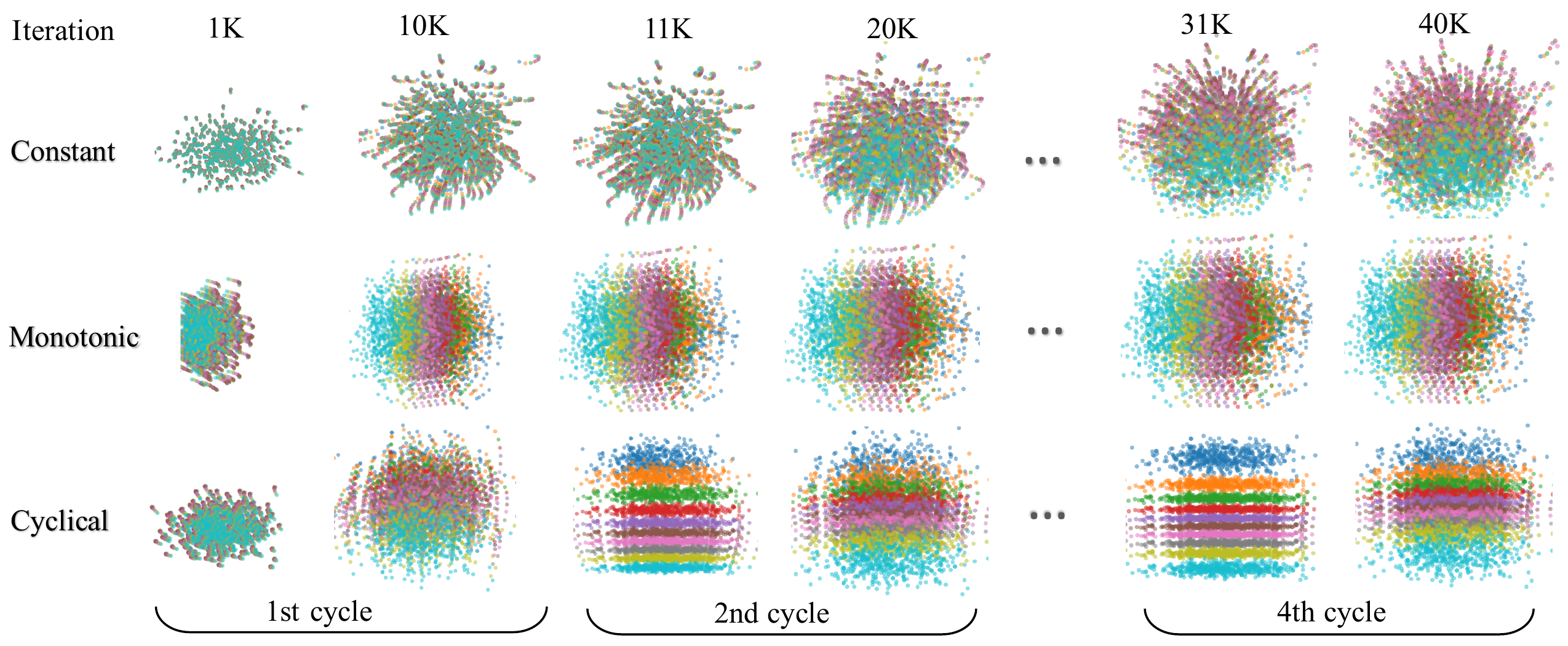}
	\vspace{-1mm}
	\caption{Visualization of the latent space along the learning dynamics on an illustrative problem.}
	\vspace{-0mm}
	\label{fig:vis_toy}
\end{figure*}

\vspace{-0mm}
\section{Visualization of Latent Space}
\vspace{-0mm}
We compare different schedule methods by visualizing the learning processes on an illustrative problem.
%To further understand different schedules, we visualize the learning dynamics on an illustrative toy problem. 
Consider a dataset consisting of 10 sequences, each of which is a 10-dimensional one-hot vector with the value 1 appearing in different positions. A 2-dimensional latent space is used for the convenience of visualization. Both the encoder and decoder are implemented using a 2-layer LSTM with 64 hidden units each. We use $T\!=\!40$K total iterations, and the scheduling schemes in Figure~\ref{fig:schedules}.

The learning curves for the ELBO, reconstruction error, and KL term are shown in Figure~\ref{fig:schedules_toy_lc}. The three schedules share very similar values. However, the cyclical schedule provides substantially lower reconstruction error and higher KL divergence. Interestingly, the cyclical schedule improves the performance progressively: it becomes better than the previous cycle, and there are clear periodic patterns across different cycles. This suggests that the cyclical schedule allows the model to use the previously learned results as a warm-restart to achieve further improvement. 
%performance.

We visualize the resulting division of the latent space for different training steps in Figure~\ref{fig:vis_toy}, where each color corresponds to $\zv \sim q(\zv|n)$, for $n=1,\cdots,10$. 
We observe that the constant schedule produces heavily mixed latent codes $\zv$ for different sequences throughout the entire training process. The monotonic schedule starts with a mixed $\zv$, but soon divides the space into a mixture of 10 cluttered Gaussians in the annealing process (the division remains cluttered in the rest of training). 
The cyclical schedule behaves similarly to the monotonic schedule in the first 10K steps (the first cycle). But, starting from the 2nd cycle, much more divided clusters are shown when learning on top of the 1st cycle results. However, $\beta<1$ leads to some holes between different clusters, making $q(\zv)$ violate the constraint of $p(\zv)$. This is alleviated at the end of the 2nd cycle, as the model is trained with $\beta=1$. As the process repeats, we see clearer patterns in the 4th cycle than the 2nd cycle for both $\beta<0$ and $\beta=1$. It shows that more structured information is captured in $\zv$ using the cyclical schedule, which is beneficial in downstream applications as shown in the experiments.

\section{Related Work}
\vspace{-0mm}

\paragraph{Solutions to KL vanishing}
Several techniques have been proposed to mitigate the KL vanishing issue. The proposed method is most closely related to the monotonic KL annealing technique in~\cite{bowman2015generating}. In addition to introducing a specific algorithm, we have  comprehensively studied the impact of $\beta$ and its scheduling schemes. Our explanations can be used to interpret other techniques, which can be broadly categorized into two classes.

The first category attempts to weaken Path B, and force the decoder to use Path A.  Word drop decoding~\cite{bowman2015generating} sets a certain percentage of the target words to zero. It has shown that it may degrade the performance when the drop rate is too high. 
The dilated CNN was considered in~\cite{yang2017improved} as a new type of decoder to replace the LSTM.
By changing the decoder's dilation architecture, one can control Path B: the effective context from $x_{<t}$.

The second category of techniques improves the dependency in Path A, so that the decoder uses latent codes more easily.
Skip connections were developed in \cite{dieng2018avoiding} to shorten the paths from $\zv$ to $\xv$ in the decoder.
\citet{zhao2017learning} introduced an auxiliary loss that requires the decoder to predict the bag-of-words in the dialog response \cite{zhao2017learning}. The decoder is thus forced to capture global information about the target response.
~\citet{zhao2017infovae} enhanced Path A via mutual information.
Concurrent with our work, \citet{he2019lagging} proposed to update encoder multiple times to achieve better latent code before updating decoder. 
Semi-amortized training \cite{kim2018semi} was proposed to perform stochastic variational inference (SVI) \cite{hoffman2013stochastic} on top of the amortized inference in VAE. It shares a similar motivation with the proposed approach, in that better latent codes can reduce KL vanishing. However, the computational cost to run SVI is high, while our monotonic schedule does not require any additional compute overhead. 
The KL scheduling methods are complementary to these techniques. As shown in experiments, the proposed cyclical schedule can further improve them.

\paragraph{$\beta$-VAE} The VAE has been extended to $\beta$-regularized versions in a growing body of work~\cite{higgins2017beta,alemi2018fixing}. Perhaps the seminal work is $\beta$-VAE~\cite{higgins2017beta}, which was extended in ~\cite{kim2018disentangling,chenisolating} to consider $\beta$ on the refined terms in the KL decomposition.
Their primary goal is to learn disentangled latent representations to explain the data, by setting $\beta>1$.
From an information-theoretic point of view, \citep{alemi2018fixing} suggests a simple method to set $\beta<1$ to ensure that latent-variable models with powerful stochastic decoders do not ignore their latent code.
However, $\beta \neq 1$ results in an improper statistical model. 
Further, $\beta$ is static in their work; we consider dynamically scheduled $\beta$ and find it more effective.

\paragraph{Cyclical schedules}
Warm-restart techniques are common in optimization to deal with multimodal functions. The cyclical schedule has been used to train deep neural networks~\cite{smith2017cyclical}, warm restart stochastic gradient descent~\cite{loshchilov2017sgdr}, improve convergence rates~\cite{smith2017super}, obtain model ensembles~\cite{huang2017snapshot} and explore multimodal distributions in MCMC sampling~\cite{zhang2019cyclical}.
All these works applied cyclical schedules to the learning rate. In contrast, this paper represents the first to consider the cyclical schedule for $\beta$ in VAE. Though the techniques seem simple and similar, our motivation is different: we use the cyclical schedule to re-open Path A in Figure~\ref{fig:schemes}(b) and provide the opportunity to train the decoder with high-quality $\zv$.   

%----------------------------------------------------------------
%----------------------------------------------------------------
%
\begin{figure*}[t!]%\vspace{-25pt}
	\vspace{-0mm}\centering
	\begin{tabular}{ccc}
	    \hspace{-3mm}		
		\includegraphics[height=3.7cm]{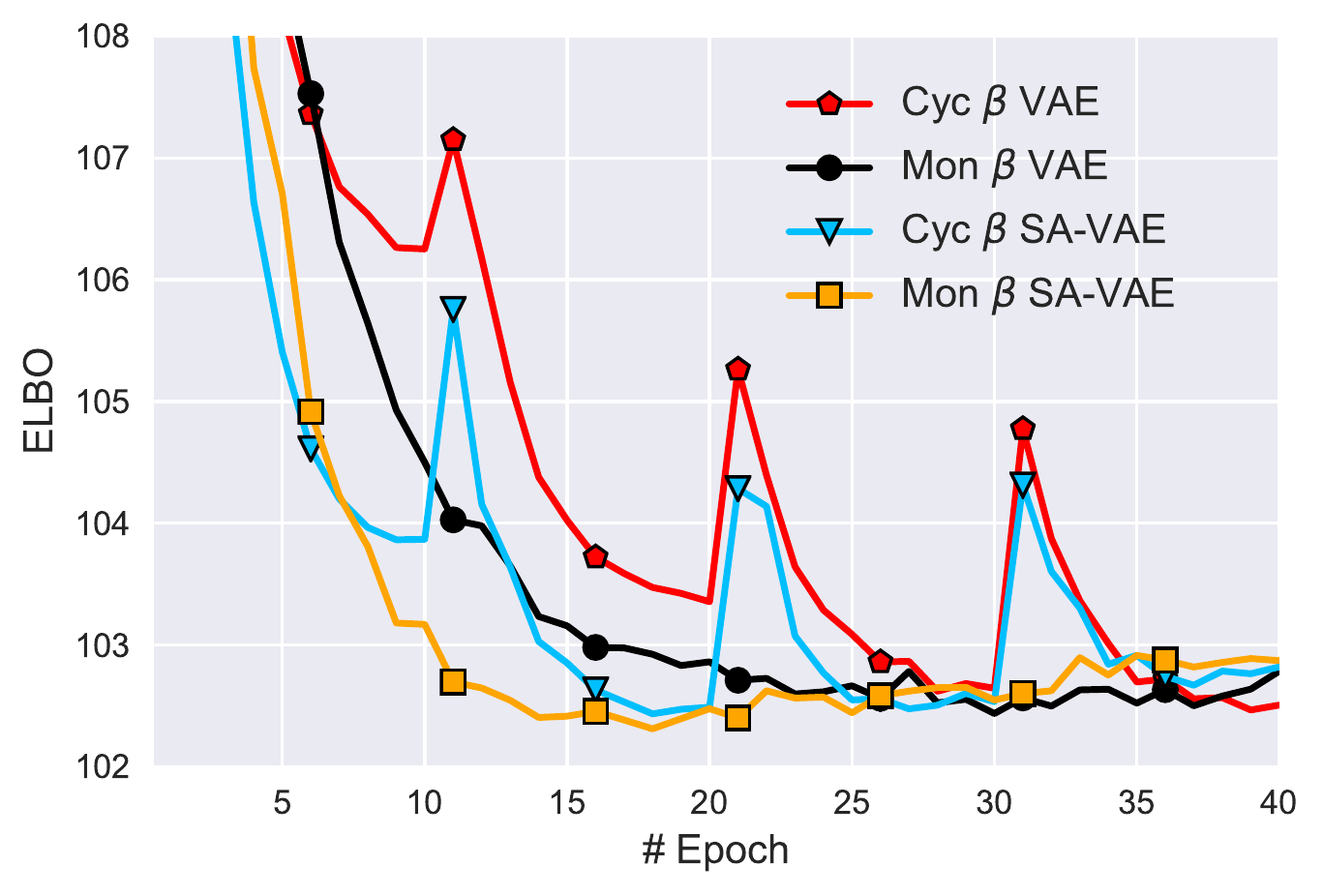}&
		\hspace{-6mm}	
		\includegraphics[height=3.7cm]{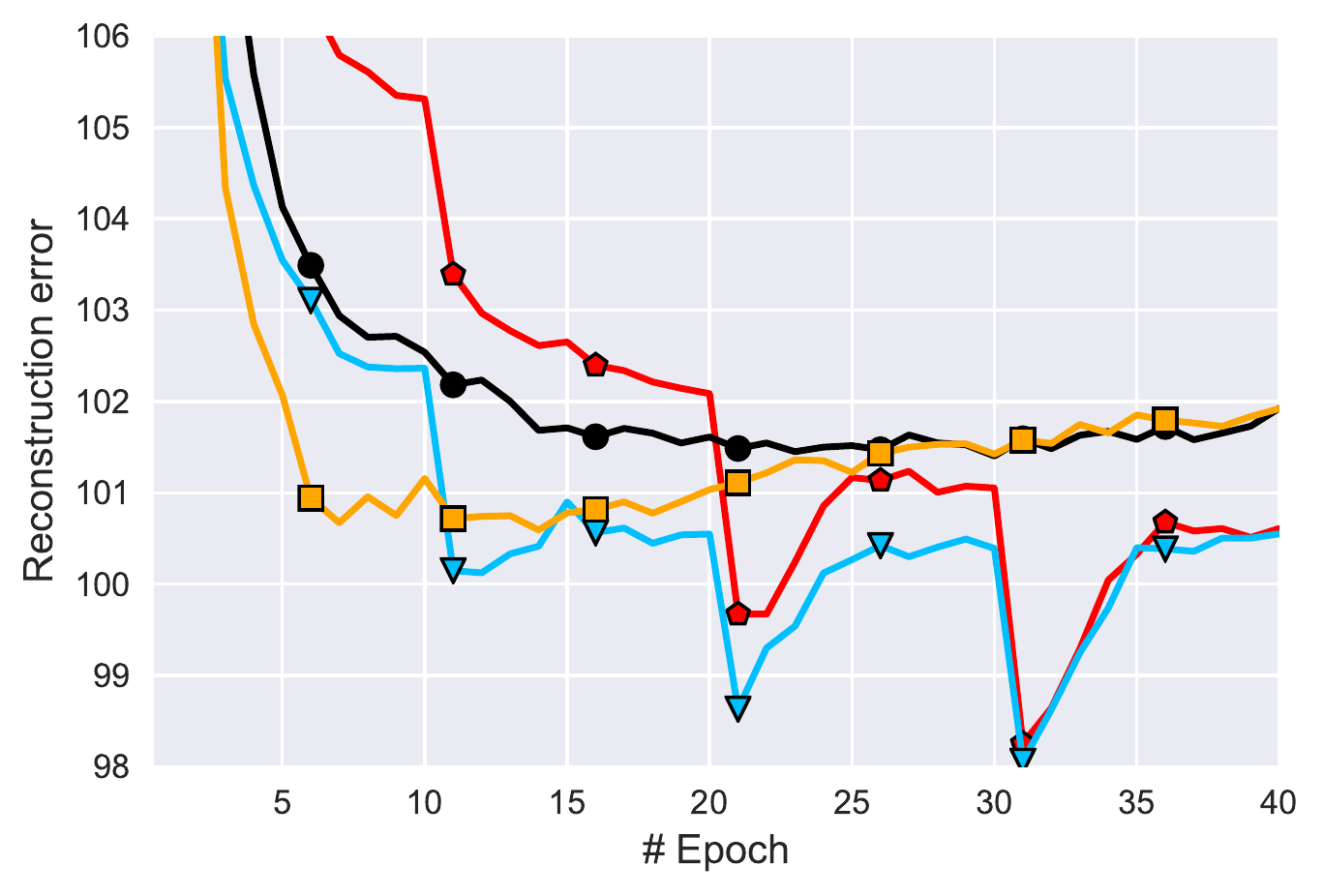} &
		\hspace{-6mm}	
		\includegraphics[height=3.7cm]{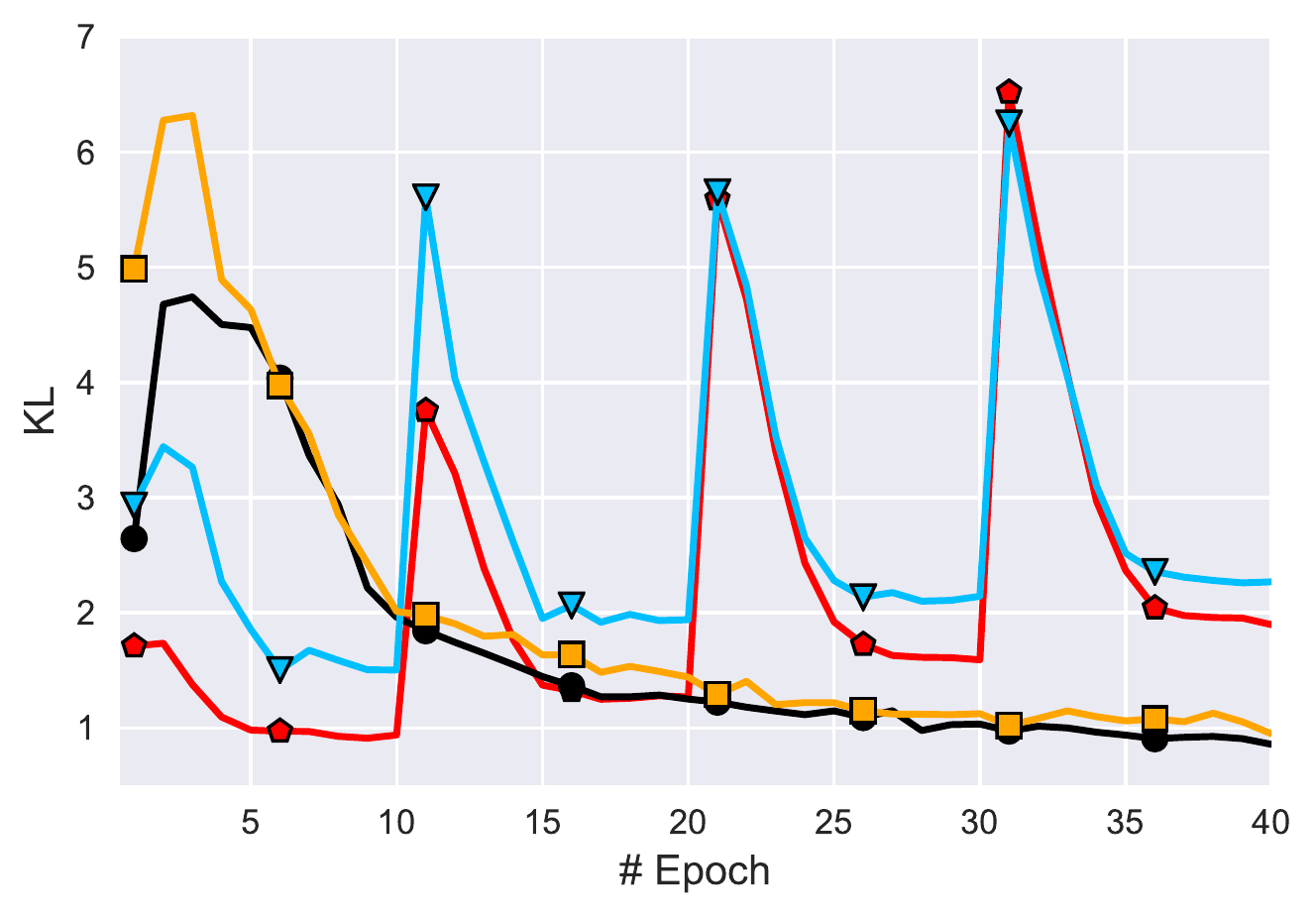} \\
		\hspace{-3mm}		
		(a) ELBO &
		(b) Reconstruction error &
		(c) KL
	\end{tabular}
	\vspace{-2mm}
	\caption{Learning curves of VAE and SA-VAE on PTB. Under similar ELBO, the cyclical schedule provides lower reconstruction errors and higher KL values than the monotonic schedule.}
	\vspace{-0mm}
	\label{fig:cyclical_schedule_ptb_supp}
\end{figure*}

\vspace{-0mm}
\section{Experiments}
\vspace{-0mm}
The source code to reproduce the experimental results will be made publicly available on GitHub\footnote{\url{https://github.com/haofuml/cyclical_annealing}}. For a fair comparison, we follow the practical recipe described in Section 3.2, where the monotonic schedule is treated as a special case of cyclical schedule (while keeping all other settings the same). The default hyper-parameters of the cyclical schedule are used in all cases unless stated otherwise. We study the impact of hyper-parameters in the SM, and show that larger $M$ can provide higher performance for various $R$. We show the major results in this section, and put more details in the SM.
The monotonic and cyclical schedules are denoted as \textbf{M} and \textbf{C}, respectively.

\subsection{Language Modeling}
%
% \JG{Is the task different from language modeling? Do you need to report perplexity in Table 1? It is useful to also present a few example sentences that are generated using the models trained with different schedules.}

We first consider language modeling on the Penn Tree Bank (PTB) dataset~\cite{marcus1993building}.
% and Yahoo questions corpus from~\cite{yang2017improved}. 
Language modeling with VAEs has been a challenging problem, and few approaches have been shown to produce rich generative models that do not collapse to standard language models. Ideally a deep generative model trained with variational inference would pursue higher ELBO, making use of the latent space (\ie maintain a nonzero KL term) while accurately modeling the underlying distribution (\ie lower reconstruction errors). We implemented different schedules based on the code\footnote{\url{https://github.com/harvardnlp/sa-vae}} published by~\citet{kim2018semi}. 

\begin{table}[t!]\centering
	\vspace{0mm}
	% \begin{adjustbox}{scale=0.9,tabular=l|l|cccc,center}
	\begin{tabular}{c}
	\hspace{-4mm}
	\includegraphics[width=8.0cm]{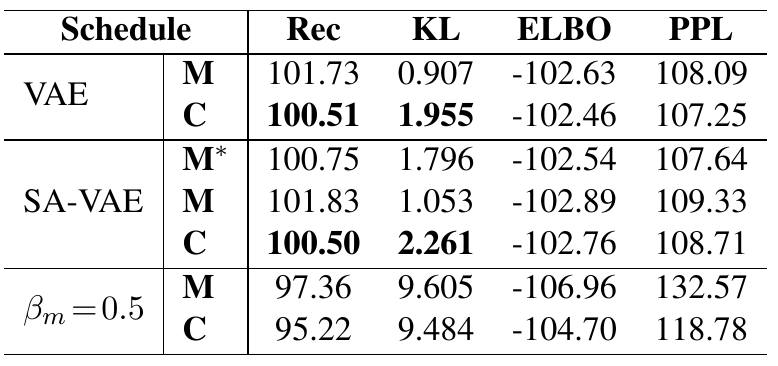}  
	\end{tabular}
	% \end{adjustbox}
	\caption{Comparison of language modeling on Penn Tree Bank (PTB). The last iteration results are reported. Since SA-VAE tends to overfit, we report its best results in row \textbf{M}$^*$.}
	\label{tab:lm}
	\vspace{-0mm}
\end{table}
%
%\begin{table}[t!]\centering
%%
%\vspace{2mm}
%% \begin{adjustbox}{scale=0.9,tabular=l|l|cccc,center}
%\begin{tabular}{l|l|cccc}
%% \begin{tabular}{l|l|c|c|c|c}
%\hline
%\multicolumn{2}{c|}{
%\textbf{Schedule}} 
%& \textbf{Rec} & \textbf{KL} & \textbf{ELBO} & \textbf{PPL} \\
%\hline
%%
%\multirow{2}{*}{VAE}
%& \textbf{M}  & 101.73 & 0.907 & -102.63  & 108.09   \\ % 108.8
%& \textbf{C}  & \textbf{100.51} & \textbf{1.955} &  -102.46 & 107.25 \\  %  
%\hline 
%\multirow{3}{*}{SA-VAE}
%& \textbf{M}$^*$ & 100.75 & 1.796 & -102.54  & 107.64   \\ % 108.8
%& \textbf{M}  & 101.83 & 1.053 &   -102.89 & 109.33 \\ % 102.33
%& \textbf{C}  & \textbf{100.50} & \textbf{2.261} &  -102.76 & 108.71 \\ % 108.2
%\hline  
%% Mon $\beta$ + Const $\eta$ &  101.96 & 0.63 & 102.59 \\ 
%%  Cyc $\beta$ + Const $\eta$ & 101.30 & 1.457 & -102.76  \\
%% Mon $\beta$ + Cyc $\eta$ & 101.74 & 0.748 & -102.49 \\
%% Cyc $\beta$ + Cyc $\eta$ &  \textbf{99.71} & \textbf{2.75} & 102.46  \\
%\multirow{2}{*}{$\beta_{m}\!=\!0.5$} 
%& \textbf{M}  & 97.36 & 9.605 & -106.96 & 132.57 \\
%& \textbf{C}  & 95.22 & 9.484 & -104.70  & 118.78   \\ % 108.2
%% Mon $\beta_{max}=0.8$  & 95.758 & 9.5607 & 105.32 \\
%\hline
%\end{tabular}
%% \end{adjustbox}
%\caption{Comparison of language modeling on Penn Tree Bank (PTB). The last iteration results are reported. Since SA-VAE tends to overfit, we report its best results in row \textbf{M}$^*$.}
%\label{tab:lm}
%\vspace{-0mm}
%\end{table}

\begin{table*}[t!]\centering
	\begin{minipage}{16.4cm}\vspace{0mm}	\centering
		% \begin{adjustbox}{scale=0.87,tabular=ll|ll,center}
		%
		\begin{tabular}{c}
		\hspace{-2mm}
		\includegraphics[width=16.0cm]{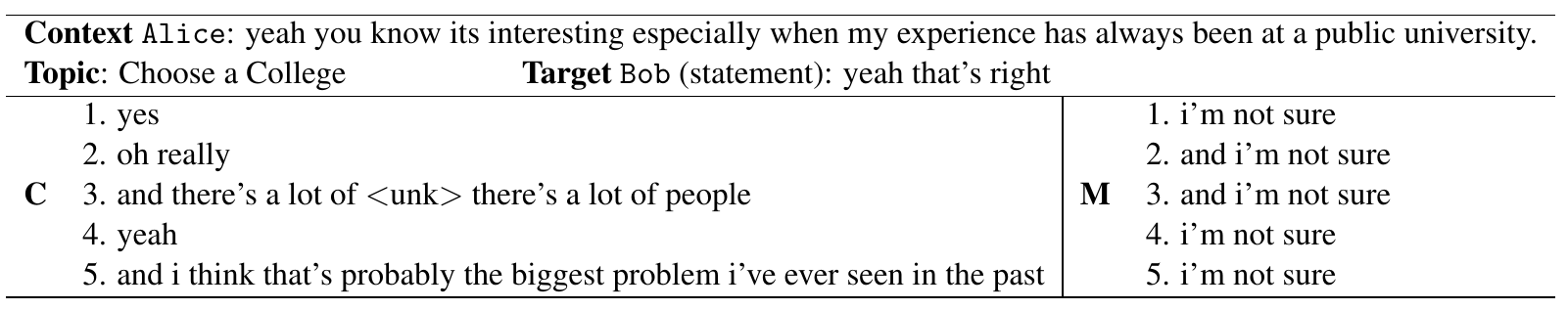}  
		\end{tabular}
		\vspace{-3mm}
		\caption{Generated dialog responses from the cyclical and monotonic schedules.}
		\label{tab:dialog_samples}
	\end{minipage}
	\vspace{-3mm}
\end{table*}

The latent variable is 32-dimensional, and 40 epochs are used.
We compare the proposed cyclical annealing schedule with the monotonic schedule baseline that, following~\cite{bowman2015generating}, anneals linearly from 0 to 1.0 over 10 epochs.
We also compare with semi-amortized (SA) training~\cite{kim2018semi}, which is considered as the state-of-the-art technique in preventing KL vanishing. We set SVI steps to 10.

Results are shown in Table~\ref{tab:lm}. The perplexity is reported in column PPL. The cyclical schedule outperforms the monotonic schedule for both standard VAE and SA-VAE training. SA-VAE training can effectively reduce KL vanishing, it takes 472s per epoch. However, this is significantly more expensive than the standard VAE training which takes 30s per epoch.
The proposed cyclical schedule adds almost zero cost.

We show the learning curves for VAE and SA-VAE in Figure~\ref{fig:cyclical_schedule_ptb_supp}. Interestingly, the cyclical schedule exhibits periodical learning behaviours. The performance of the cyclical schedule gets better progressively, after each cycle. While ELBO and PPL ar similar, the cyclical schedule improves the reconstruction ability and KL values for both VAE and SA-VAE. We observe clear over-fitting issues for the SA-VAE with the monotonic schedule, while this issue is less severe for SA-VAE with the cyclical schedule.

Finally, we further investigate whether our improvements are from simply having a lower $\beta$, rather than from the cyclical schedule re-opening Path A for better learning. To test this, we use a monotonic schedule with maximum $\beta=0.5$. We observe that the reconstruction and KL terms perform better individually, but the ELBO is substantially worse than $\beta=1$, because $\beta=0.5$ yields an improper model. Even so, the cyclical schedule improves its performance.

\begin{table}[t!]\centering
\vspace{2mm}
\begin{tabular}{l||c|c||c|c}
\hline
\textbf{Model} & \multicolumn{2}{c||}{CVAE} & 
\multicolumn{2}{c}{CVAE+BoW}  \\ \hline
\textbf{Schedule} & \textbf{M} & \textbf{C}
& \textbf{M} & \textbf{C} \\
\hline
Rec-P $\downarrow$ & 
36.16 & \textbf{29.77} & 18.44 & \textbf{16.74} \\
KL Loss $\uparrow$ & 
0.265 & \textbf{4.104} & 14.06 & \textbf{15.55} \\ \hline
B4 prec & 
0.185 & \textbf{0.234} & 0.211 & \textbf{0.219} \\
B4 recall & 
0.122 & \textbf{0.220} & 0.210 & \textbf{0.219} \\
A-bow prec &
0.957 & \textbf{0.961} & 0.958 & \textbf{0.961} \\
A-bow recall &
0.911 & \textbf{0.941} & 0.938 & \textbf{0.940} \\
E-bow prec &
0.867 & 0.833 & 0.830 & 0.828 \\
E-bow recall &
0.784 & \textbf{0.808} & 0.808 & 0.805 \\
\hline
\end{tabular}
\caption{Comparison on dialog response generation. Reconstruction perplexity (Rec-P) and BLEU (B) scores are used for evaluation.}
\label{tab:dialog}
\vspace{-3mm}
\end{table}

\subsection{Conditional VAE for Dialog}
\label{sec_exp:dialog}
We use a cyclical schedule to improve the latent codes in~\cite{zhao2017learning}, which are key to diverse dialog-response generation. 
Following~\cite{zhao2017learning}, Switchboard (SW) Corpus~\cite{godfrey1997switchboard} is used, which has 2400 two-sided telephone conversations.

Two latent variable models are considered. The first one is the Conditional VAE (CVAE), which has been shown better than the encoder-decoder neural dialog~\cite{serban2016building}. The second is to augment VAE with a bag-of-word (BoW) loss to tackle the KL vanishing problem, as proposed in~\cite{zhao2017learning}. 

Table 2 shows the sample outputs generated from the two schedules using CVAE. Caller $\mathtt{Alice}$ begins with an open-ended statement on choosing a college, and the model learns to generate responses from Caller $\mathtt{Bob}$. The cyclical schedule generated highly diverse answers that cover multiple plausible dialog acts. On the contrary, the
responses from the monotonic schedule are limited to repeat plain responses, \ie ``{\it i'm not sure}''. 

Quantitative results are shown in Table~\ref{tab:dialog}, using the evaluation metrics from~\cite{zhao2017learning}.
$(\RN{1})$ 
Smoothed Sentence-level BLEU~\cite{chen2014systematic}: BLEU is a popular metric that measures the geometric mean of modified n-gram precision with a length penalty. We use BLEU-1 to 4 as our lexical similarity metric and normalize the score to 0 to 1 scale.
$(\RN{2})$ Cosine Distance of Bag-of-word Embedding~\cite{liu2016not}: a simple method to obtain sentence embeddings is to take the average or extreme of all the word embeddings in the sentences. We used Glove embedding and denote the average method as {\it A$-$bow} and extreme method as {\it E$-$bow}. The score is normalized to $[0, 1]$.
Higher values indicate more plausible responses. 
% The full results are in the SM.

The BoW indeed reduces the KL vanishing issue, as indicated by the increased KL and decreased reconstruction perplexity. When applying the proposed cyclical schedule to CVAE, we also see a reduced KL vanishing issue. Interestingly, it also yields the highest BLEU scores. This suggests that the cyclical schedule can generate dialog responses of higher fidelity with lower cost, as the auxiliary BoW loss is not necessary. Further, BoW can be improved when integrated with the cyclical schedule, as shown in the last column of Table~\ref{tab:dialog}.

\subsection{Unsupervised Language Pre-training}
We consider the Yelp dataset, as pre-processed in~\cite{shen2017style} for unsupervised language pre-training. Text features are extracted as the latent codes $\zv$ of VAE models, pre-trained with monotonic and cyclical schedules.
The AE is used as the baseline. A good VAE can learn to cluster data into meaningful groups~\cite{kingma2013auto}, indicating that well-structured $\zv$ are highly informative features, which usually leads to higher classification performance. To clearly compare the quality of $\zv$, we build a simple one-layer classifier on $\zv$, and fine-tune the model on different proportions of labelled data~\cite{zhang2017deconvolutional}.

\begin{figure}[t!]%\vspace{-25pt}
	\vspace{-0mm}\centering
	\includegraphics[width=5.75cm]{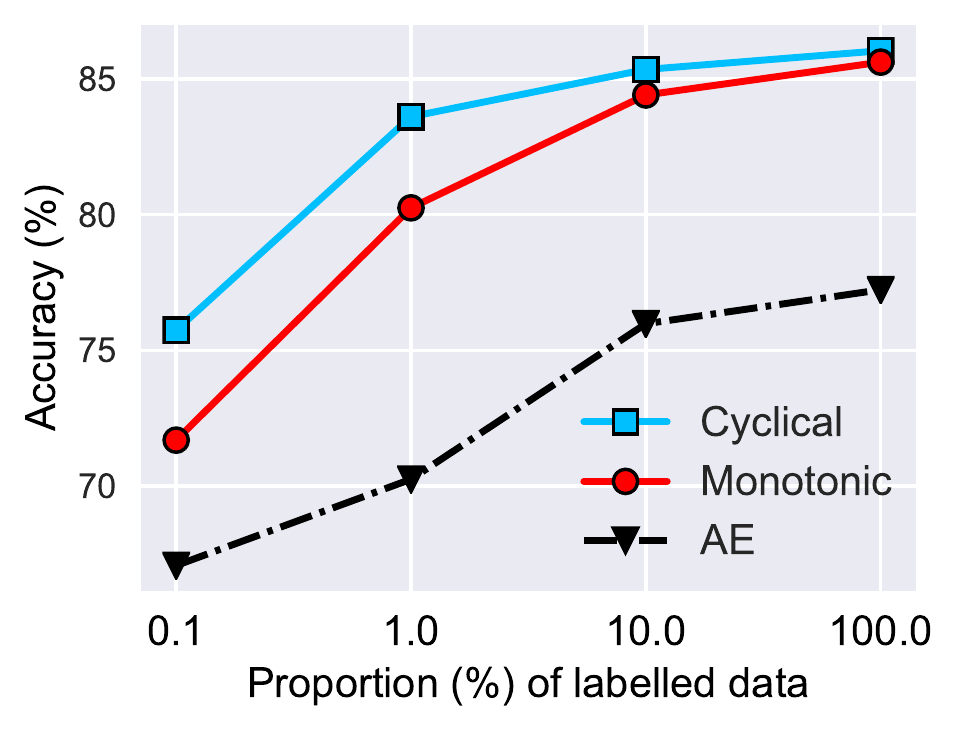}  
	\vspace{-3mm}
	\caption{\small Accuracy of fine-tuning on the unsupervised pre-trained models on the Yelp dataset.}
	\vspace{2mm}
	\label{fig:ssl_acc}
\end{figure}

The results are shown in Figure~\ref{fig:ssl_acc}. The cyclical schedule consistently yields the highest accuracy relative to other methods. We visualize the tSNE embeddings~\cite{maaten2008visualizing} of $\zv$ in Figure~\ref{fig:tsne_schedules_supp} of the SM, and observe that the cyclical schedule exhibits clearer clustered patterns.

% VAEs for Unsupervised Text Generation

% \subsection{Hyper-parameter study}

\begin{table}[t!]\centering
\begin{tabular}{l||c|c|c}
\hline
\textbf{Schedule} & \textbf{Rec} & \textbf{KL} & \textbf{ELBO} \\
\hline
Cyc $\beta$ + Const $\eta$ & \textbf{101.30} & \textbf{1.457} & -102.76  \\
Mon $\beta$ + Const $\eta$  & 101.93 & 0.858 & -102.78   \\ % 108.8
\hline 
Cyc $\beta$ + Cyc $\eta$   & \textbf{100.61} & \textbf{1.897} &  -102.51  \\  %  107.5
Mon $\beta$ + Cyc $\eta$ & 101.74 & 0.748 & -102.49 \\
\hline
\end{tabular}
\caption{Comparison of cyclical schedules on $\beta$ and $\eta$, tested with language modeling on PTB.}
\label{tab:lm_supp_ptb}
\vspace{-3mm}
\end{table}

\subsection{Ablation Study}

To enhance the performance, we propose to apply the cyclical schedule to the learning rate $\eta$ on real tasks.
It ensures that the optimizer has the same length of optimization trajectory for each $\beta$ cycle (so that each cycle can fully converge). To investigate the impact of cyclical on $\eta$, we perform two more ablation experiments:
$(\RN{1})$ We make only $\beta$ cyclical, keep $\eta$ constant. 
$(\RN{2})$ We make only $\eta$ cyclical, keep $\beta$ monotonic.
The last epoch numbers are shown in Table~\ref{tab:lm_supp_ptb}, and the learning curves on shown in Figure~\ref{fig:cyclical_eta_beta_supp} in SM.
Compared with the baseline, we see that it is the cyclical $\beta$ rather than cyclical $\eta$ that contributes to the improved performance.

\vspace{-0mm}
\section{Conclusions}
\vspace{-0mm}
We provide a novel two-path interpretation to explain the KL vanishing issue, and identify its source as a lack of good latent codes at the beginning of decoder training. This provides an understanding of various $\beta$ scheduling schemes, and motivates the proposed cyclical schedule. By re-opening the path at $\beta=0$, the cyclical schedule can progressively improve the performance, by leveraging good latent codes learned in the previous cycles as warm re-starts. 
We demonstrate the effectiveness of the proposed approach on three NLP tasks, and show that it is superior to or complementary to other techniques.

% \newpage
\section*{Acknowledgments}
We thank Yizhe Zhang, Sungjin Lee, Dinghan Shen, Wenlin Wang for insightful discussion. The implementation in our experiments heavily depends on three NLP applications published on Github repositories; we acknowledge all the authors who made their code public, which tremendously accelerates our project progress.

\bibliography{naaclhlt2019}

\begin{thebibliography}{43}
\expandafter\ifx\csname natexlab\endcsname\relax\def\natexlab#1{#1}\fi

\bibitem[{Alemi et~al.(2018)Alemi, Poole, Fischer, Dillon, Saurous, and
  Murphy}]{alemi2018fixing}
Alexander Alemi, Ben Poole, Ian Fischer, Joshua Dillon, Rif~A Saurous, and
  Kevin Murphy. 2018.
\newblock Fixing a broken {ELBO}.
\newblock In \emph{ICML}.

\bibitem[{Bowman et~al.(2015)Bowman, Vilnis, Vinyals, Dai, Jozefowicz, and
  Bengio}]{bowman2015generating}
Samuel~R Bowman, Luke Vilnis, Oriol Vinyals, Andrew~M Dai, Rafal Jozefowicz,
  and Samy Bengio. 2015.
\newblock Generating sentences from a continuous space.
\newblock \emph{arXiv preprint arXiv:1511.06349}.

\bibitem[{Chen and Cherry(2014)}]{chen2014systematic}
Boxing Chen and Colin Cherry. 2014.
\newblock A systematic comparison of smoothing techniques for sentence-level
  bleu.
\newblock In \emph{Proceedings of the Ninth Workshop on Statistical Machine
  Translation}, pages 362--367.

\bibitem[{Chen et~al.(2018)Chen, Li, Grosse, and Duvenaud}]{chenisolating}
Ricky~TQ Chen, Xuechen Li, Roger Grosse, and David Duvenaud. 2018.
\newblock Isolating sources of disentanglement in {VAE}s.
\newblock \emph{NIPS}.

\bibitem[{Dieng et~al.(2018)Dieng, Kim, Rush, and Blei}]{dieng2018avoiding}
Adji~B Dieng, Yoon Kim, Alexander~M Rush, and David~M Blei. 2018.
\newblock Avoiding latent variable collapse with generative skip models.
\newblock \emph{arXiv preprint arXiv:1807.04863}.

\bibitem[{Fraccaro et~al.(2016)Fraccaro, S{\o}nderby, Paquet, and
  Winther}]{fraccaro2016sequential}
Marco Fraccaro, S{\o}ren~Kaae S{\o}nderby, Ulrich Paquet, and Ole Winther.
  2016.
\newblock Sequential neural models with stochastic layers.
\newblock In \emph{NIPS}.

\bibitem[{Godfrey and Holliman(1997)}]{godfrey1997switchboard}
J~Godfrey and E~Holliman. 1997.
\newblock Switchboard-1 release 2: Linguistic data consortium.
\newblock \emph{SWITCHBOARD: A User's Manual}.

\bibitem[{Goodfellow et~al.(2016)Goodfellow, Bengio, and
  Courville}]{goodfellow2016deep}
Ian Goodfellow, Yoshua Bengio, and Aaron Courville. 2016.
\newblock \emph{Deep learning}, volume~1.
\newblock MIT press Cambridge.

\bibitem[{Goyal et~al.(2017)Goyal, Sordoni, C{\^o}t{\'e}, Ke, and
  Bengio}]{goyal2017z}
Anirudh Goyal Alias~Parth Goyal, Alessandro Sordoni, Marc-Alexandre
  C{\^o}t{\'e}, Nan~Rosemary Ke, and Yoshua Bengio. 2017.
\newblock Z-forcing: Training stochastic recurrent networks.
\newblock In \emph{NIPS}.

\bibitem[{He et~al.(2019)He, Spokoyny, Neubig, and
  Berg-Kirkpatrick}]{he2019lagging}
Junxian He, Daniel Spokoyny, Graham Neubig, and Taylor Berg-Kirkpatrick. 2019.
\newblock Lagging inference networks and posterior collapse in variational
  autoencoders.
\newblock \emph{ICLR}.

\bibitem[{Higgins et~al.(2017)Higgins, Matthey, Pal, Burgess, Glorot,
  Botvinick, Mohamed, and Lerchner}]{higgins2017beta}
Irina Higgins, Loic Matthey, Arka Pal, Christopher Burgess, Xavier Glorot,
  Matthew Botvinick, Shakir Mohamed, and Alexander Lerchner. 2017.
\newblock beta-vae: Learning basic visual concepts with a constrained
  variational framework.
\newblock \emph{ICLR}.

\bibitem[{Hochreiter and Schmidhuber(1997)}]{hochreiter1997long}
Sepp Hochreiter and Jurgen Schmidhuber. 1997.
\newblock Long short-term memory.
\newblock \emph{Neural computation}.

\bibitem[{Hoffman et~al.(2013)Hoffman, Blei, Wang, and
  Paisley}]{hoffman2013stochastic}
Matthew~D Hoffman, David~M Blei, Chong Wang, and John Paisley. 2013.
\newblock Stochastic variational inference.
\newblock \emph{The Journal of Machine Learning Research}.

\bibitem[{Hoffman and Johnson(2016)}]{hoffman2016elbo}
Matthew~D Hoffman and Matthew~J Johnson. 2016.
\newblock Elbo surgery: yet another way to carve up the variational evidence
  lower bound.
\newblock In \emph{Workshop in Advances in Approximate Bayesian Inference,
  NIPS}.

\bibitem[{Hu et~al.(2017)Hu, Yang, Liang, Salakhutdinov, and
  Xing}]{hu2017toward}
Zhiting Hu, Zichao Yang, Xiaodan Liang, Ruslan Salakhutdinov, and Eric~P Xing.
  2017.
\newblock Toward controlled generation of text.
\newblock \emph{ICML}.

\bibitem[{Huang et~al.(2017)Huang, Li, Pleiss, Liu, Hopcroft, and
  Weinberger}]{huang2017snapshot}
Gao Huang, Yixuan Li, Geoff Pleiss, Zhuang Liu, John~E Hopcroft, and Kilian~Q
  Weinberger. 2017.
\newblock Snapshot ensembles: Train 1, get m for free.
\newblock \emph{ICLR}.

\bibitem[{Kim and Mnih(2018)}]{kim2018disentangling}
Hyunjik Kim and Andriy Mnih. 2018.
\newblock Disentangling by factorising.
\newblock \emph{ICML}.

\bibitem[{Kim et~al.(2018)Kim, Wiseman, Miller, Sontag, and Rush}]{kim2018semi}
Yoon Kim, Sam Wiseman, Andrew~C Miller, David Sontag, and Alexander~M Rush.
  2018.
\newblock Semi-amortized variational autoencoders.
\newblock \emph{ICML}.

\bibitem[{Kingma and Welling(2013)}]{kingma2013auto}
Diederik~P Kingma and Max Welling. 2013.
\newblock Auto-encoding variational bayes.
\newblock \emph{ICLR}.

\bibitem[{Lai et~al.(2018)Lai, Li, Zheng, and Yang}]{lai2018stochastic}
Guokun Lai, Bohan Li, Guoqing Zheng, and Yiming Yang. 2018.
\newblock Stochastic wavenet: A generative latent variable model for sequential
  data.
\newblock \emph{ICML workshop}.

\bibitem[{Li et~al.(2017)Li, Liu, Chen, Pu, Chen, Henao, and
  Carin}]{li2017alice}
Chunyuan Li, Hao Liu, Changyou Chen, Yuchen Pu, Liqun Chen, Ricardo Henao, and
  Lawrence Carin. 2017.
\newblock {ALICE}: Towards understanding adversarial learning for joint
  distribution matching.
\newblock In \emph{NIPS}.

\bibitem[{Liu et~al.(2016)Liu, Lowe, Serban, Noseworthy, Charlin, and
  Pineau}]{liu2016not}
Chia-Wei Liu, Ryan Lowe, Iulian~V Serban, Michael Noseworthy, Laurent Charlin,
  and Joelle Pineau. 2016.
\newblock How not to evaluate your dialogue system: An empirical study of
  unsupervised evaluation metrics for dialogue response generation.
\newblock \emph{arXiv preprint arXiv:1603.08023}.

\bibitem[{Loshchilov and Hutter(2017)}]{loshchilov2017sgdr}
Ilya Loshchilov and Frank Hutter. 2017.
\newblock Sgdr: Stochastic gradient descent with warm restarts.
\newblock \emph{ICLR}.

\bibitem[{Maaten and Hinton(2008)}]{maaten2008visualizing}
Laurens van~der Maaten and Geoffrey Hinton. 2008.
\newblock Visualizing data using t-sne.
\newblock \emph{Journal of machine learning research}, 9(Nov):2579--2605.

\bibitem[{Makhzani et~al.(2016)Makhzani, Shlens, Jaitly, Goodfellow, and
  Frey}]{makhzani2016adversarial}
Alireza Makhzani, Jonathon Shlens, Navdeep Jaitly, Ian Goodfellow, and Brendan
  Frey. 2016.
\newblock Adversarial autoencoders.
\newblock \emph{ICLR workshop}.

\bibitem[{Marcus et~al.(1993)Marcus, Marcinkiewicz, and
  Santorini}]{marcus1993building}
Mitchell~P Marcus, Mary~Ann Marcinkiewicz, and Beatrice Santorini. 1993.
\newblock Building a large annotated corpus of english: The penn treebank.
\newblock \emph{Computational linguistics}.

\bibitem[{Miao and Blunsom(2016)}]{miao2016language}
Yishu Miao and Phil Blunsom. 2016.
\newblock Language as a latent variable: Discrete generative models for
  sentence compression.
\newblock \emph{EMNLP}.

\bibitem[{Miao et~al.(2016)Miao, Yu, and Blunsom}]{miao2016neural}
Yishu Miao, Lei Yu, and Phil Blunsom. 2016.
\newblock Neural variational inference for text processing.
\newblock In \emph{ICML}.

\bibitem[{Mikolov et~al.(2010)Mikolov, Karafi{\'a}t, Burget,
  {\v{C}}ernock{\`y}, and Khudanpur}]{mikolov2010recurrent}
Tom{\'a}{\v{s}} Mikolov, Martin Karafi{\'a}t, Luk{\'a}{\v{s}} Burget, Jan
  {\v{C}}ernock{\`y}, and Sanjeev Khudanpur. 2010.
\newblock Recurrent neural network based language model.
\newblock In \emph{Eleventh Annual Conference of the International Speech
  Communication Association}.

\bibitem[{Rezende et~al.(2014)Rezende, Mohamed, and
  Wierstra}]{rezende2014stochastic}
Danilo~Jimenez Rezende, Shakir Mohamed, and Daan Wierstra. 2014.
\newblock Stochastic backpropagation and approximate inference in deep
  generative models.
\newblock \emph{ICML}.

\bibitem[{Serban et~al.(2016)Serban, Sordoni, Bengio, Courville, and
  Pineau}]{serban2016building}
Iulian~Vlad Serban, Alessandro Sordoni, Yoshua Bengio, Aaron~C Courville, and
  Joelle Pineau. 2016.
\newblock Building end-to-end dialogue systems using generative hierarchical
  neural network models.
\newblock In \emph{AAAI}.

\bibitem[{Shen et~al.(2017)Shen, Lei, Barzilay, and Jaakkola}]{shen2017style}
Tianxiao Shen, Tao Lei, Regina Barzilay, and Tommi Jaakkola. 2017.
\newblock Style transfer from non-parallel text by cross-alignment.
\newblock In \emph{NIPS}.

\bibitem[{Shu et~al.(2018)Shu, Bui, Zhao, Kochenderfer, and
  Ermon}]{shu2018amortized}
Rui Shu, Hung~H Bui, Shengjia Zhao, Mykel~J Kochenderfer, and Stefano Ermon.
  2018.
\newblock Amortized inference regularization.
\newblock \emph{NIPS}.

\bibitem[{Smith(2017)}]{smith2017cyclical}
Leslie~N Smith. 2017.
\newblock Cyclical learning rates for training neural networks.
\newblock In \emph{WACV}. IEEE.

\bibitem[{Smith and Topin(2017)}]{smith2017super}
Leslie~N Smith and Nicholay Topin. 2017.
\newblock Super-convergence: Very fast training of residual networks using
  large learning rates.
\newblock \emph{arXiv preprint arXiv:1708.07120}.

\bibitem[{Wen et~al.(2017)Wen, Miao, Blunsom, and Young}]{wen2017latent}
Tsung-Hsien Wen, Yishu Miao, Phil Blunsom, and Steve Young. 2017.
\newblock Latent intention dialogue models.
\newblock \emph{ICML}.

\bibitem[{Xu et~al.(2017)Xu, Sun, Deng, and Tan}]{xu2017variational}
Weidi Xu, Haoze Sun, Chao Deng, and Ying Tan. 2017.
\newblock Variational autoencoder for semi-supervised text classification.
\newblock In \emph{AAAI}.

\bibitem[{Yang et~al.(2017)Yang, Hu, Salakhutdinov, and
  Berg-Kirkpatrick}]{yang2017improved}
Zichao Yang, Zhiting Hu, Ruslan Salakhutdinov, and Taylor Berg-Kirkpatrick.
  2017.
\newblock Improved variational autoencoders for text modeling using dilated
  convolutions.
\newblock \emph{ICML}.

\bibitem[{Zhang et~al.(2017{\natexlab{a}})Zhang, Bengio, Hardt, Recht, and
  Vinyals}]{zhang2017understanding}
Chiyuan Zhang, Samy Bengio, Moritz Hardt, Benjamin Recht, and Oriol Vinyals.
  2017{\natexlab{a}}.
\newblock Understanding deep learning requires rethinking generalization.
\newblock \emph{ICLR}.

\bibitem[{Zhang et~al.(2019)Zhang, Li, Zhang, Chen, and
  Wilson}]{zhang2019cyclical}
Ruqi Zhang, Chunyuan Li, Jianyi Zhang, Changyou Chen, and Andrew~Gordon Wilson.
  2019.
\newblock Cyclical stochastic gradient mcmc for bayesian deep learning.
\newblock \emph{arXiv preprint arXiv:1902.03932}.

\bibitem[{Zhang et~al.(2017{\natexlab{b}})Zhang, Shen, Wang, Gan, Henao, and
  Carin}]{zhang2017deconvolutional}
Yizhe Zhang, Dinghan Shen, Guoyin Wang, Zhe Gan, Ricardo Henao, and Lawrence
  Carin. 2017{\natexlab{b}}.
\newblock Deconvolutional paragraph representation learning.
\newblock In \emph{NIPS}.

\bibitem[{Zhao et~al.(2019)Zhao, Song, and Ermon}]{zhao2017infovae}
Shengjia Zhao, Jiaming Song, and Stefano Ermon. 2019.
\newblock Info{VAE}: Information maximizing variational autoencoders.
\newblock \emph{AAAI}.

\bibitem[{Zhao et~al.(2017)Zhao, Zhao, and Eskenazi}]{zhao2017learning}
Tiancheng Zhao, Ran Zhao, and Maxine Eskenazi. 2017.
\newblock Learning discourse-level diversity for neural dialog models using
  conditional variational autoencoders.
\newblock \emph{ACL}.

\end{thebibliography}
\bibliographystyle{acl_natbib}

\newpage
~
\newpage
\appendix

\section{Comparison of different schedules}
\label{sec_supp:schedules}
\begin{figure*}[t!]%\vspace{-25pt}
	\vspace{-0mm}\centering
	\begin{tabular}{ccc}
	    \hspace{-3mm}		
		\includegraphics[height=3.4cm]{figs/schedules/beta_iteration_linear2.pdf} &
		\includegraphics[height=3.4cm]{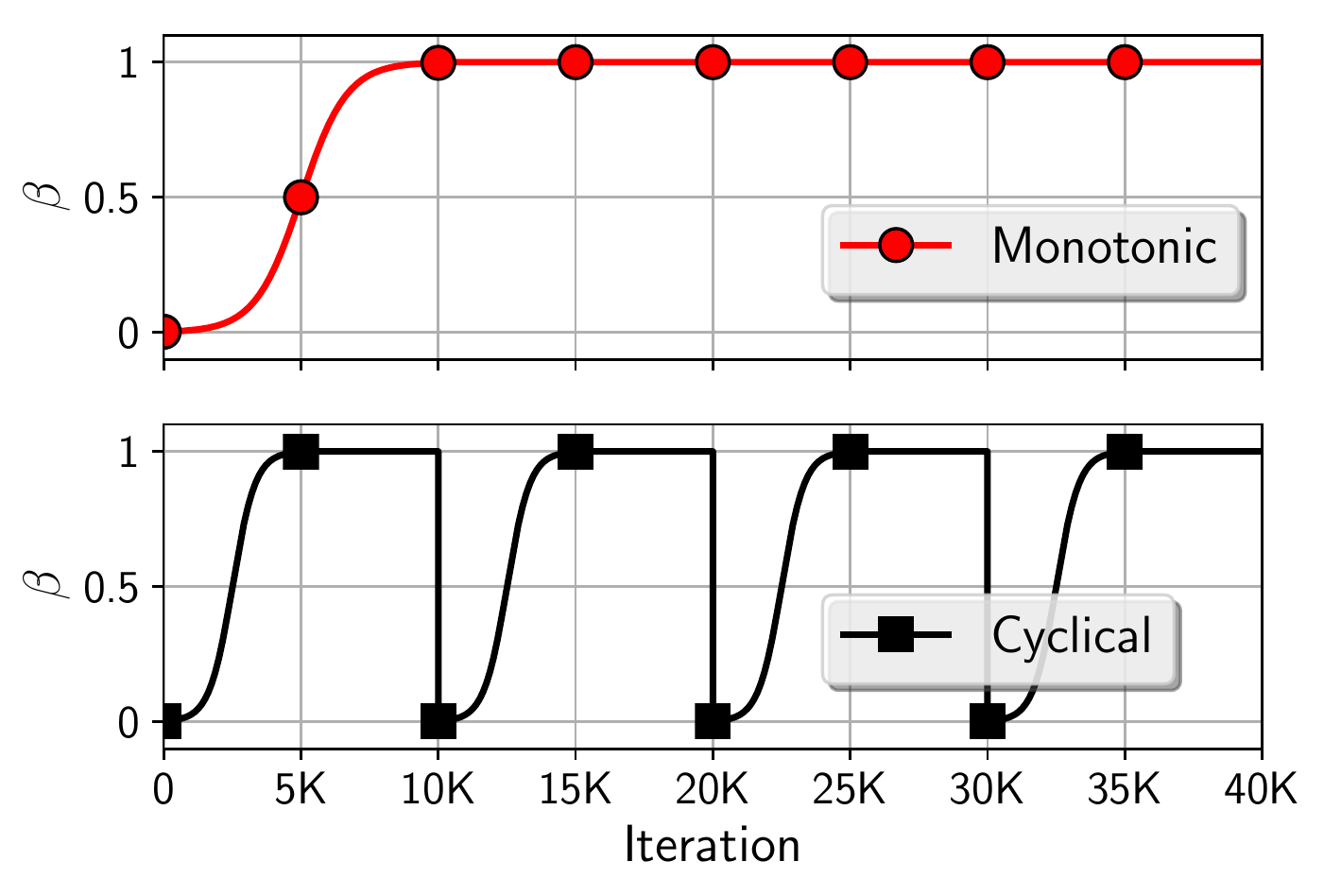} &
		\includegraphics[height=3.4cm]{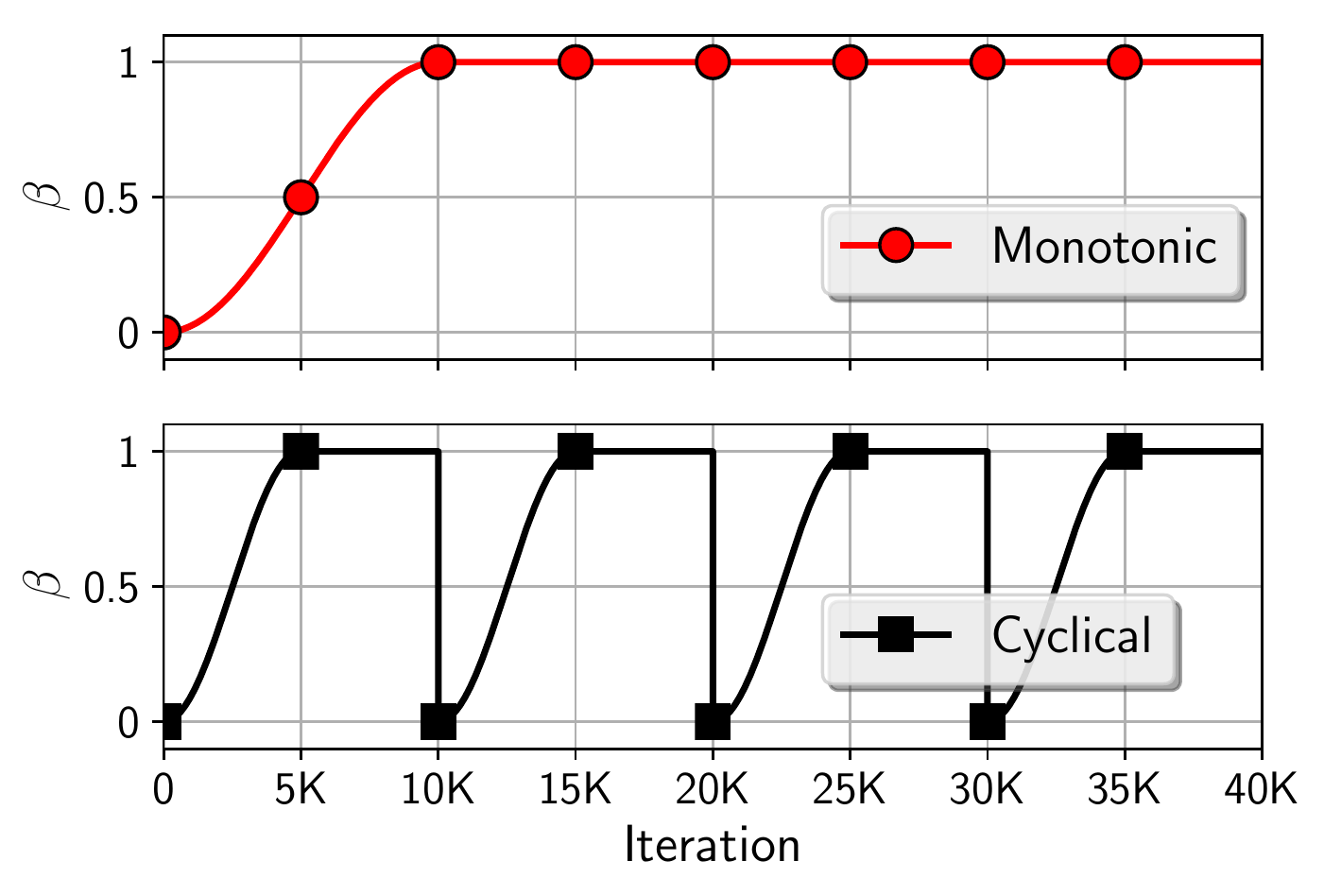} \\
		\hspace{-3mm}		
		(a) Linear &
		(b) Sigmoid &
		(c) Cosine
	\end{tabular}
	\vspace{-2mm}
	\caption{Comparison between traditional monotonic and proposed cyclical annealing schedules. The top row shows the traditional monotonic schedules, and the bottom row shows their corresponding cyclical schedules. $M=4$ cycles are illustrated, $R=0.5$ is used for annealing within each cycle.}
	\vspace{-2mm}
	\label{fig:schedules_supp}
\end{figure*}

We compare the two different scheduling schemes in Figure~\ref{fig:schedules_supp}. The three widely used monotonic schedules are shown in the top row, including linear, sigmoid and cosine. We can easily turn them into their corresponding cyclical versions, shown in the bottom row.

\section{Proofs on the $\beta$ and MI}
When scheduled with $\beta$, the training objective over the dataset can be written as:
\begin{align} \label{eq_ml_analysis_supp}
\Fcal & = -\Fcal_E + \beta \Fcal_R 
\end{align}
We proceed the proof by re-writing each term separately.

\subsection{Bound on $\Fcal_E$ }
Following~\cite{li2017alice}, on the support of $(\xv, \zv)$, we denote $q$ as the encoder probability measure, and $p$ as the decoder probability measure. Note that the reconstruction loss for $\zv$ can be writen as its negative log likelihood form as:  
\begin{align} \label{eq_rec_analysis_supp}
\Fcal_E = -\E_{\xv\sim q(\xv), \zv \sim q(\zv|\xv)}[\log p(\xv|\zv)].
\end{align}

\begin{lemma}\label{lemma:infogan}
	For random variables $\xv$ and $\zv$ with two different probability measures, $p(\xv, \zv)$ and $q(\xv, \zv)$, we have
	\begin{align}
	& H_{p}(\zv|\xv) \nonumber \\
	& = - \E_{\zv \sim p(\zv), x\sim p(\xv|\zv)}[\log p(\zv|\xv)] \nonumber \\
	&= -\E_{\zv \sim p(\zv), \xv \sim p( \xv |\zv)}[\log q(\zv | \xv)] \nonumber \\
	& - \E_{ \zv \sim p( \zv), x\sim p( \xv| \zv )}\big[\log p( \zv | \xv ) - \log q( \zv | \xv )\big] \nonumber \\
	&= - \E_{\zv \sim p( \zv ), \xv \sim p( \xv |\zv)}[\log q(\zv | \xv )] \nonumber  \\
	& - \E_{p( \xv )}(\mbox{KL}(p( \zv | \xv )\| q( \zv | \xv )))
	\nonumber \\
	&\le -\E_{\zv \sim p(\zv), \xv \sim p(\xv | \zv )}[\log q( \zv | \xv)]
	\end{align}
\end{lemma}	
where $H_p(\zv|\xv)$ is the conditional entropy. Similarly, we can prove that 
\begin{align}
H_{q}(\xv|\zv) \le -\E_{\xv \sim q(\xv), \zv \sim q(\zv | \xv )}[\log p( \xv | \zv)]
\end{align}

From lemma \ref{lemma:infogan}, we have 
\begin{corollary}
	For random variables $\xv$ and $\zv$ with  probability measure $p(\xv, \zv)$, the mutual information between $\xv$ and $\zv$ can be written as
	\begin{align}
	I_q(\xv, \zv) 
	& =  H_q(\xv) - H_q(\xv|\zv) \ge H_q(\xv) \nonumber \\
	& + \E_{\xv \sim q(\xv), \zv \sim q(\zv|\xv)}[\log p(\xv|\zv)]
	\nonumber \\
	& = H_q(\xv) + \Fcal_E
	\end{align}	
\end{corollary}

\subsection{Decomposition of $\Fcal_R$}
The KL term in~\eqref{eq_reg_elbo} can be decomposed into two refined terms~\cite{hoffman2016elbo}: 
\vspace{-3mm}
\begin{align} \label{eq_kl_decomp_supp}
& \Fcal_R \nonumber \\
& = \E_{q(\xv)}[ \mbox{KL} (q(\zv | \xv) || p(\zv) ) ]  \nonumber \\
& = \E_{q(\zv, \xv)} [ (\log q(\zv | \xv) - \log p(\zv) ) ]  \nonumber \\
& = \E_{q(\zv, \xv)} [ (\log q(\zv | \xv) - \log q(\zv)]   \nonumber  \\
& \hspace{10mm}  + \E_{q(\zv, \xv)} [\log q(\zv) - \log p(\zv) ) ]  \nonumber \\
& = \E_{q(\zv, \xv)} [ (\log q(\zv , \xv) - \log q(\xv) - \log q(\zv)] \nonumber \\
& \hspace{10mm} + \E_{q(\zv, \xv)} [\log q(\zv)  - \log p(\zv) ) ]  \nonumber \\
& =  \underbrace{I_q(\zv, \xv)}_{\Fcal_1:~\text{Mutual~Info.}} +
\underbrace{ \mbox{KL}(q(\zv) || p(\zv)) }_{\Fcal_2:~\text{Marginal KL} } 
\end{align} 

\section{Model Description}

\subsection{Conditional VAE for dialog}
Each conversation can be represented via three random variables: the dialog context $\cv$ composed of the dialog history, the response utterance $\xv$, and a latent variable $\zv$, which is used to capture the latent distribution over the valid responses($\beta=1$)
 \cite{zhao2017learning}.
The ELBO can be written as:
\begin{align}
\log p_{\thetav}(\xv|\cv) & \ge \Lcal_{ \text{ELBO} }  \\
&  =
\E_{q_{\phiv}(\zv | \xv, \cv)} \big[ \log p_{\thetav}(\xv | \zv, \cv) \big] \nonumber \\
&  - \beta \mbox{KL} (q_{\phiv}(\zv | \xv, \cv) || p(\zv| \cv) ) \nonumber
%\label{eq_vae_elbo}
\end{align}
 
\subsection{Semi-supervised learning with VAE}
We use a simple factorization to derive the ELBO for semi-supervised learning. $\alpha$ is introduced to regularize the strength of classification loss.
\begin{align}
& \log p_{\thetav}(\yv, \xv)  \ge \Lcal_{ \text{ELBO} }  \\
&  =
\E_{q_{\phiv}(\zv | \xv)} \big[ 
\log p_{\thetav}(\xv | \zv)  + 
\alpha \log p_{\psiv}(\yv | \zv) 
\big] \nonumber \\
&  - \beta \mbox{KL} (q_{\phiv}(\zv | \xv ) || p(\zv) )
%\label{eq_vae_elbo}
\end{align}
where $\psiv$ is the parameters for the classifier.

Good latent codes $\zv$ are crucial for the the classification performance, especially when simple classifiers are employed, or less labelled data is available.

\section{More Experimental Results}

%\subsection{Language Modeling on PTB}
%\paragraph{Code $\&$ Evaluation}
% We implemented different schedules based on the code\footnote{\url{https://github.com/harvardnlp/sa-vae}} published by~\citet{kim2018semi}. 
% %
% Note that ELBO is closely related to PPL. 
% $\mbox{PPL} = \ell/N_s$ and $\mbox{PPL} = \exp(\ell/N_w)$, where $\ell$ is the total loss, $N_s$ is the number of sentences, and $N_w$ is the number of words. 

% \paragraph{Results}  We show the learning curves for VAE and SA-VAE in Figure~\ref{fig:cyclical_schedule_ptb_supp}. The cyclical schedule exhibits periodical learning behaviours. While ELBO and PPL ar similar, the cyclical schedule improves the reconstruction ability and KL values.

\subsection{CVAE for Dialog Response Generation}
\paragraph{Code $\&$ Dataset}
We implemented different schedules based on the code\footnote{\url{https://github.com/snakeztc/NeuralDialog-CVAE}} published by~\citet{zhao2017learning}. 
In the SW dataset, there are 70 available topics. We randomly split the data into 2316/60/62 dialogs for train/validate/test.

\begin{figure*}[t!]%\vspace{-25pt}
	\vspace{-0mm}\centering
	\begin{tabular}{cc}
	    \hspace{-3mm}		
		\includegraphics[height=4.4cm]{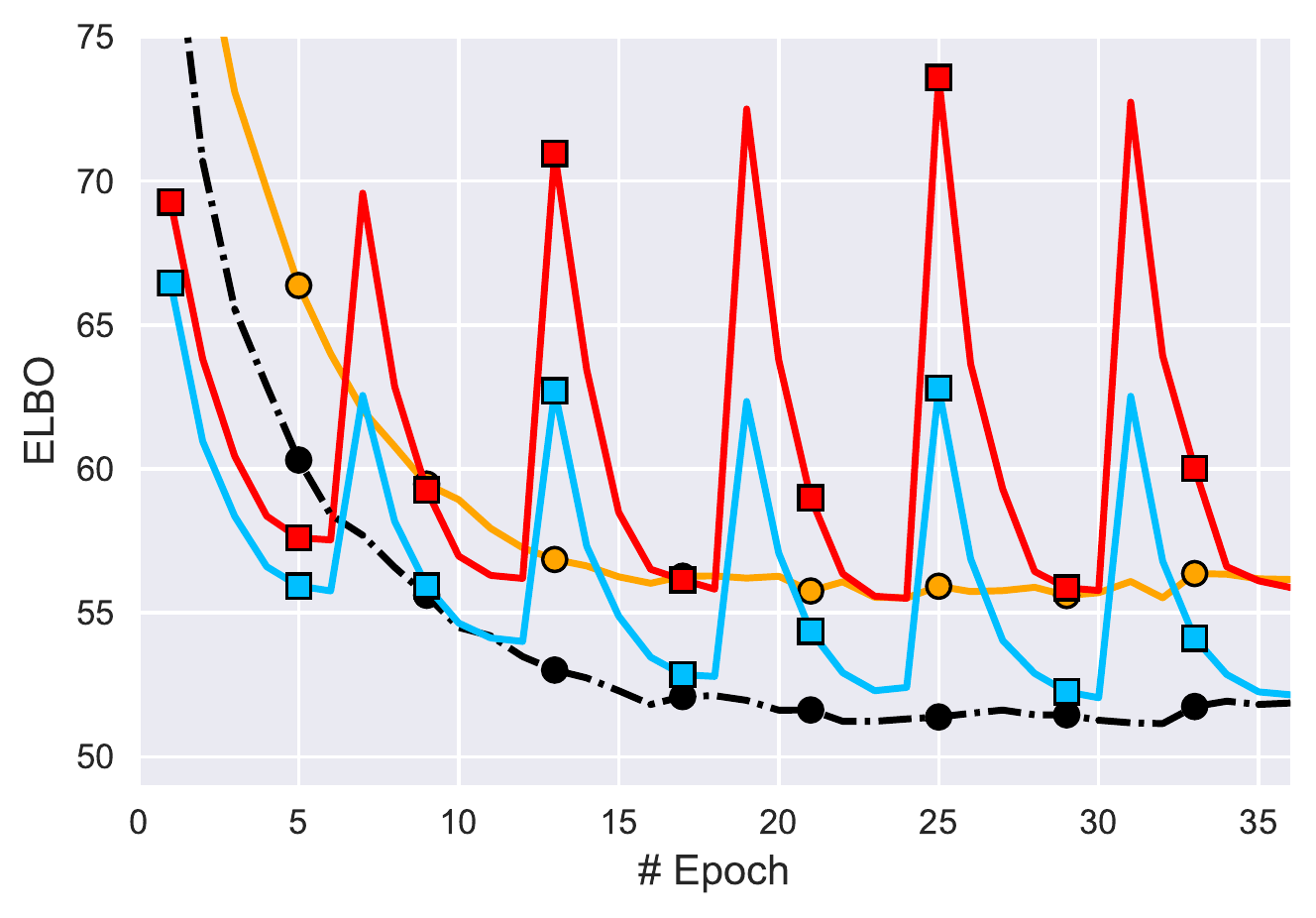} &
		\includegraphics[height=4.4cm]{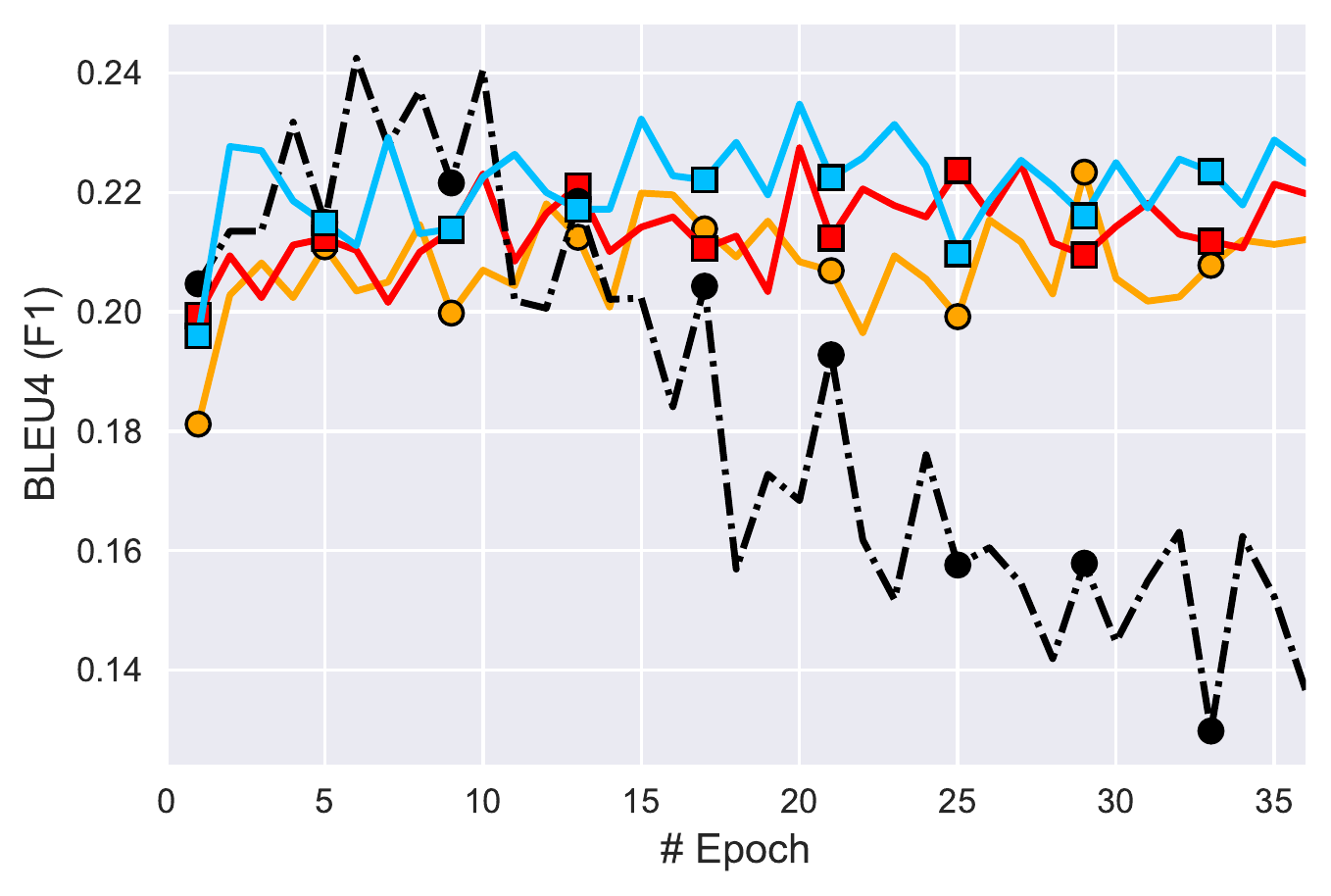} \\
		(a) ELBO &
		(b) BLEU-4 (F1)  \\
		\includegraphics[height=4.4cm]{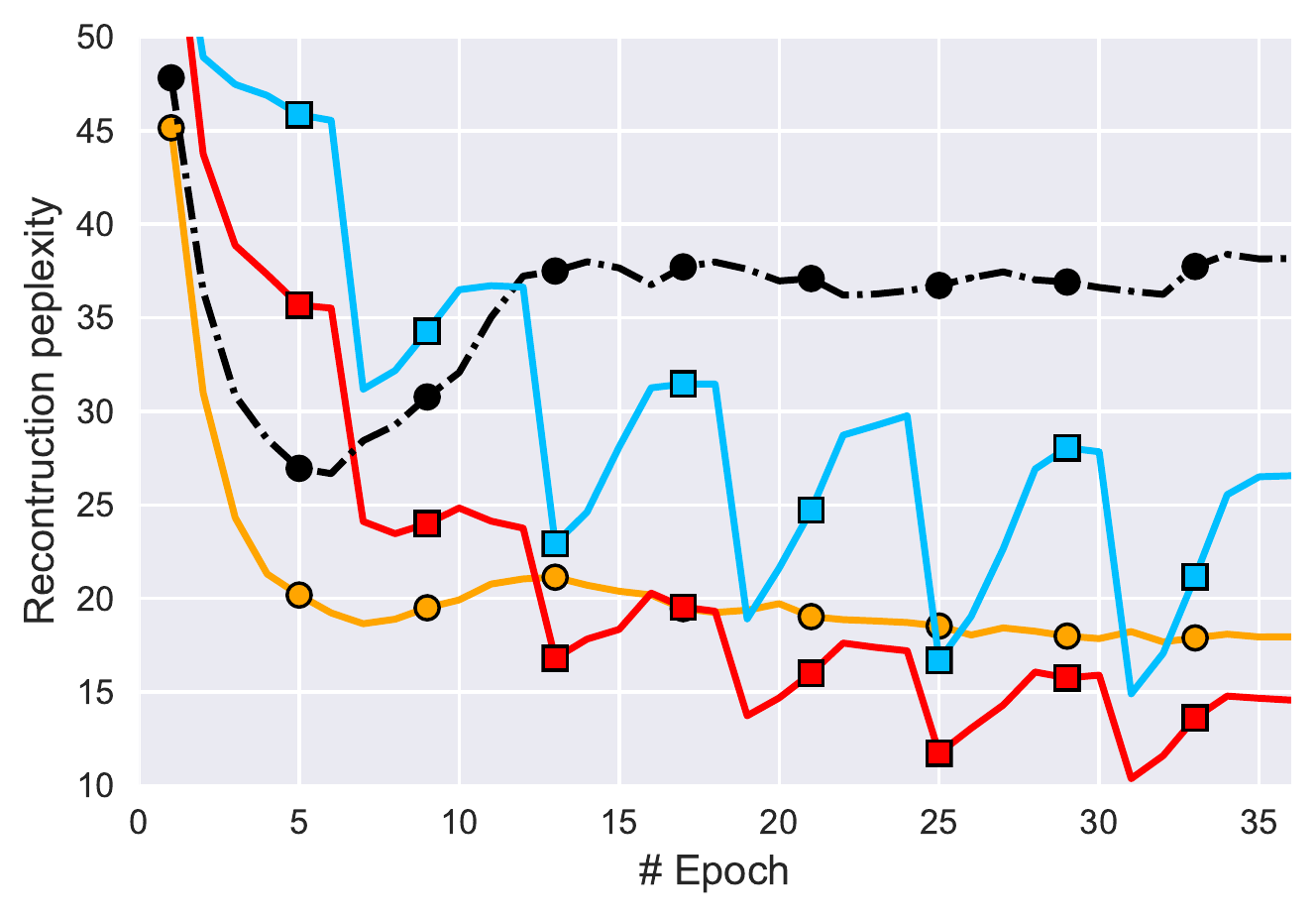} &		
		\includegraphics[height=4.4cm]{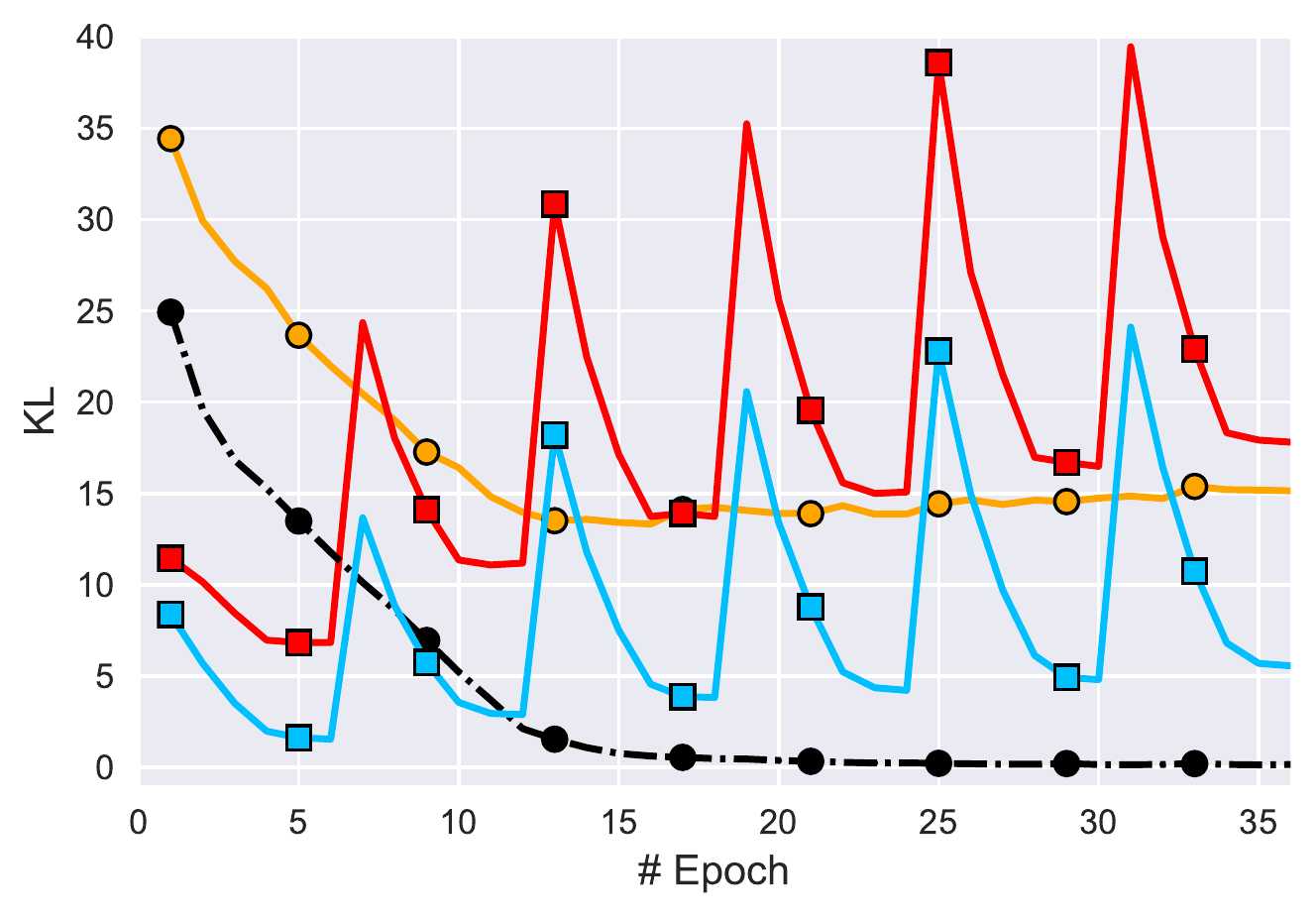} \\
		\hspace{-3mm}		
		(c) Reconstruction error &
		(d) KL \\
		&
		\hspace{-90mm}
		\frame{
		\includegraphics[height=0.4cm]{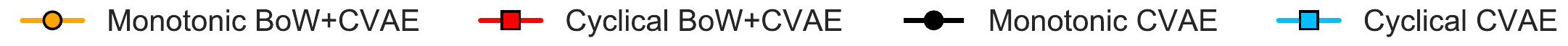}
		}
		 \hspace{-20mm}
		
	\end{tabular}
	\vspace{-2mm}
	\caption{Full results of CVAE and BoW+CVAE on SW dataset. Under similar ELBO results, the cyclical schedule provide lower reconstruction errors, higher KL values, and higher BLEU values than the monotonic schedule. Interestingly, the monotonic schedule tends to overfit, while the cyclical schedule does not.}
	\vspace{-2mm}
	\label{fig:cyclical_schedule_sw_supp}
\end{figure*}

\paragraph{Results}  The results on full BLEU scores are shown in Table~\ref{tab:dialog_supp}. The cyclical schedule outperforms the monotonic schedule in both settings.
The learning curves are shown in Figure~\ref{fig:cyclical_schedule_sw_supp}.
Under similar ELBO results, the cyclical schedule provide lower reconstruction errors, higher KL values, and higher BLEU values than the monotonic schedule. Interestingly, the monotonic schedule tends to overfit, while the cyclical schedule does not, particularly on reconstruction errors. It means the monotonic schedule can learn better latent codes for VAEs, thus preventing overfitting.

\subsection{Semi-supervised Text Classification}
\paragraph{Dataset}
Yelp restaurant reviews dataset utilizes user ratings associated with each review. Reviews with rating above three are considered positive, and those below three are considered negative. 
Hence, this is a binary classification problem. 
The pre-processing in~\cite{shen2017style} allows sentiment analysis on sentence level. It further filters the sentences by eliminating those that exceed 15 words. The resulting dataset has 250K negative sentences, and 350K positive ones. The vocabulary size is 10K after replacing words occurring less than 5 times with the ``$<$unk$>$'' token.

\begin{figure*}[t!]%\vspace{-25pt}
	\vspace{-0mm}\centering
	\begin{tabular}{ccc}
	    \hspace{-3mm}		
		\includegraphics[height=3.9cm]{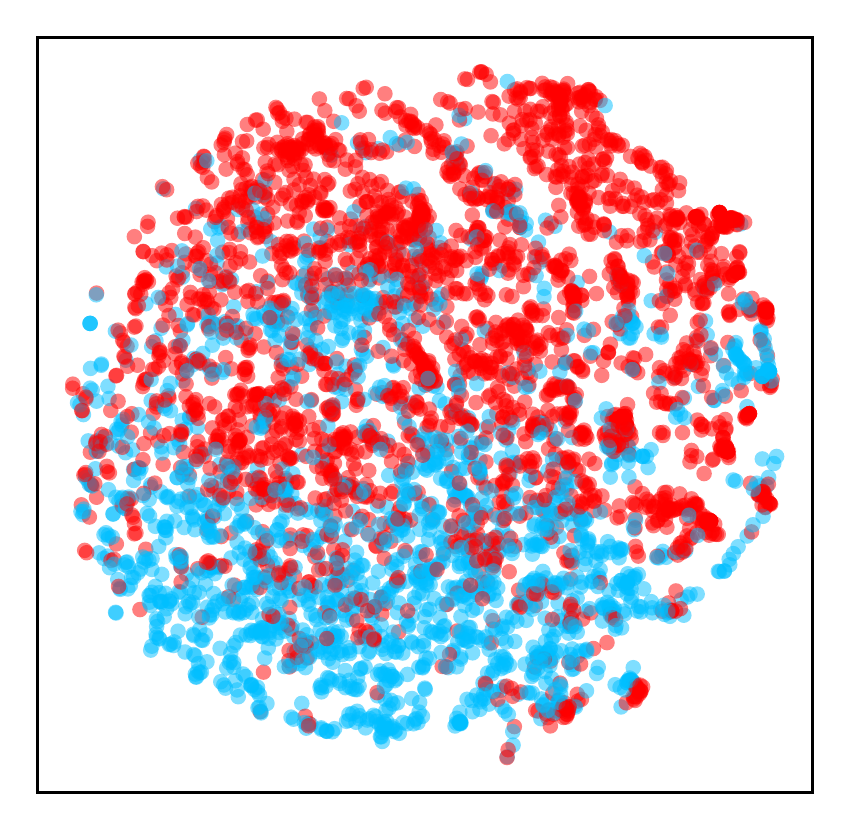} &
		\includegraphics[height=3.9cm]{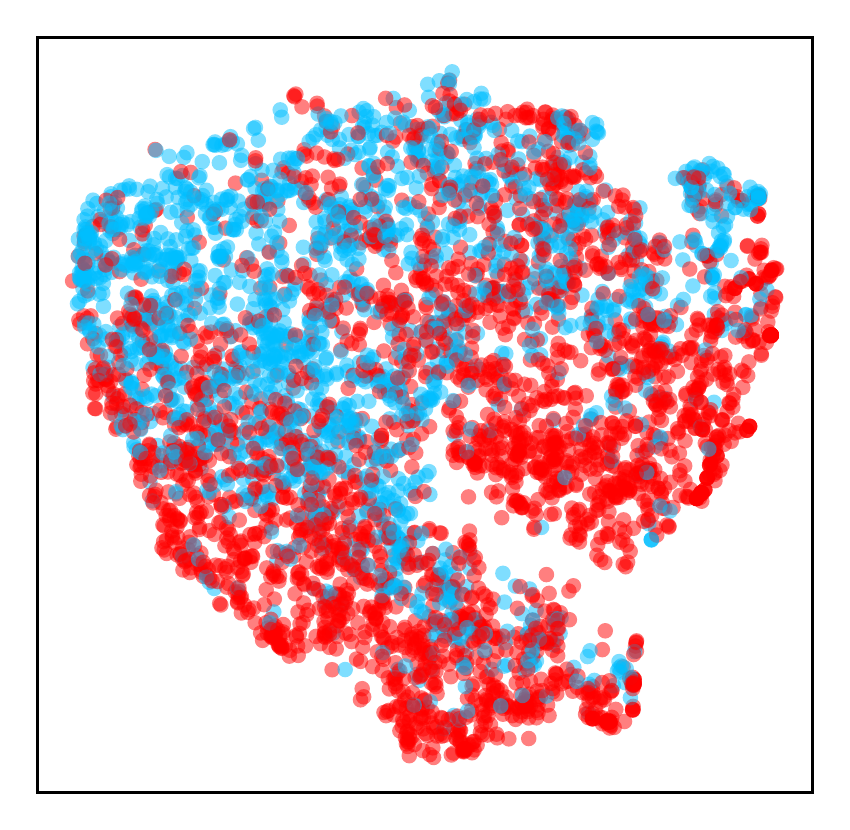} &
		\includegraphics[height=3.9cm]{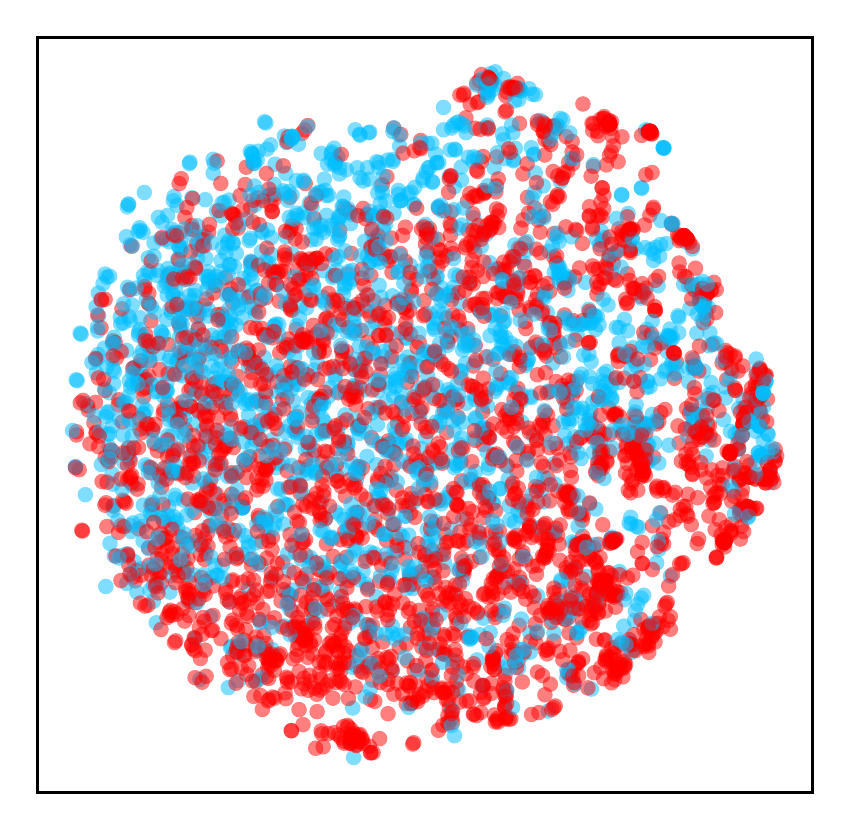} \\
		\hspace{-3mm}		
		(a) Cyclical $\beta$ VAE &
		(b) Monotonic $\beta$ VAE  &
		(c) AE
	\end{tabular}
	\vspace{-2mm}
	\caption{Comparison of tSNE embeddings for three methods on Yelp dataset. This can be considered as the unsupervised feature learning results in semi-supervised learning. More structured latent patterns usually lead to better classification performance.}
	\vspace{-2mm}
	\label{fig:tsne_schedules_supp}
\end{figure*}

\paragraph{Results} The tSNE embeddings are visualized in Figure~\ref{fig:tsne_schedules_supp}. We see that cyclical $\beta$ provides much more separated latent structures than the other two methods.

\subsection{Hyper-parameter tuning}
\begin{figure*}[t!]%\vspace{-25pt}
	\vspace{-0mm}\centering
	\begin{tabular}{ccc}
	    \hspace{-3mm}		
		\includegraphics[height=3.6cm]{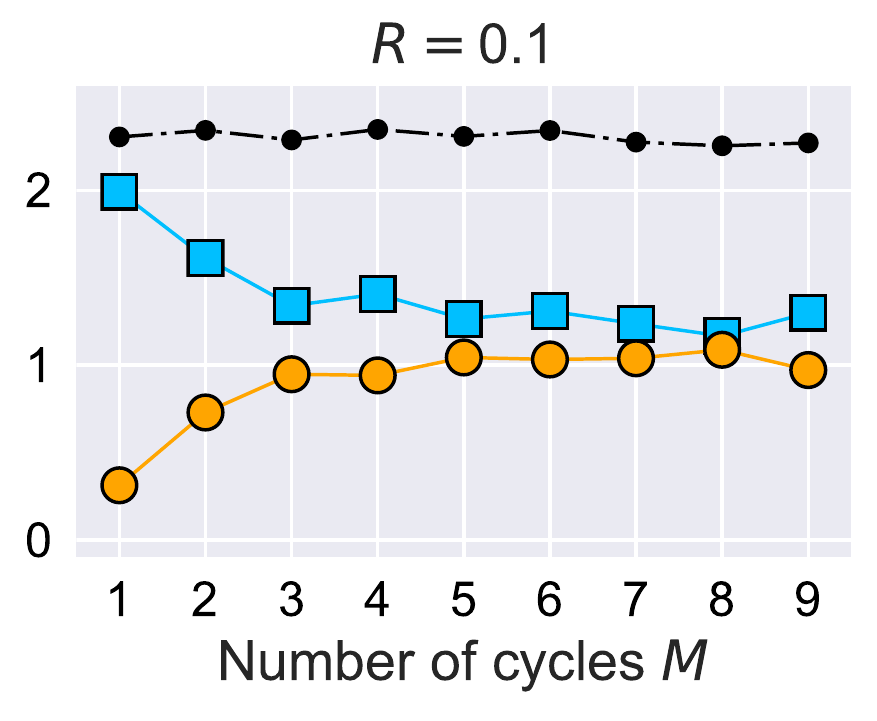} &
		\includegraphics[height=3.6cm]{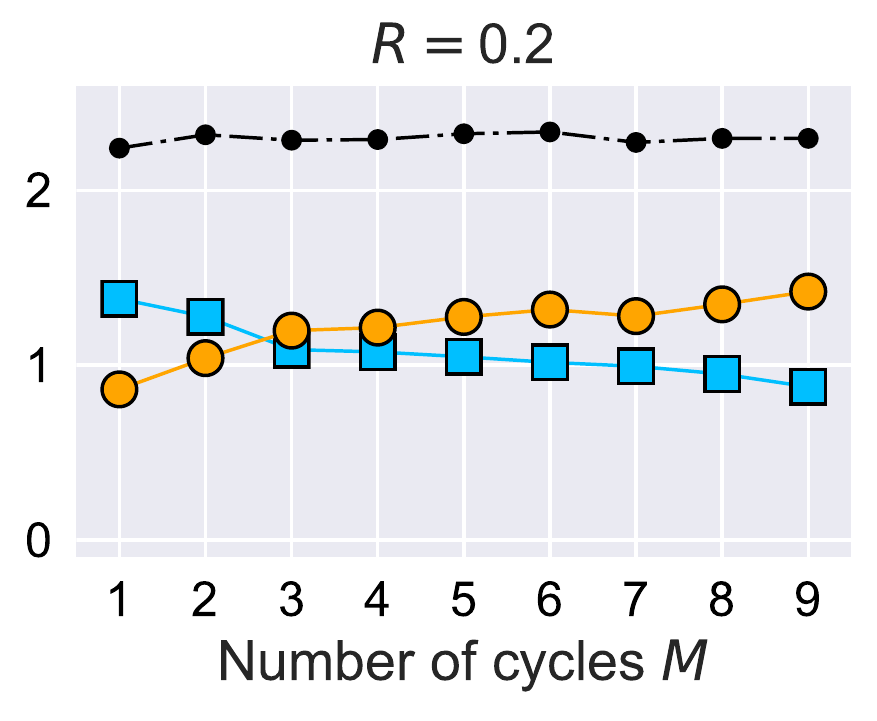} &
		\includegraphics[height=3.6cm]{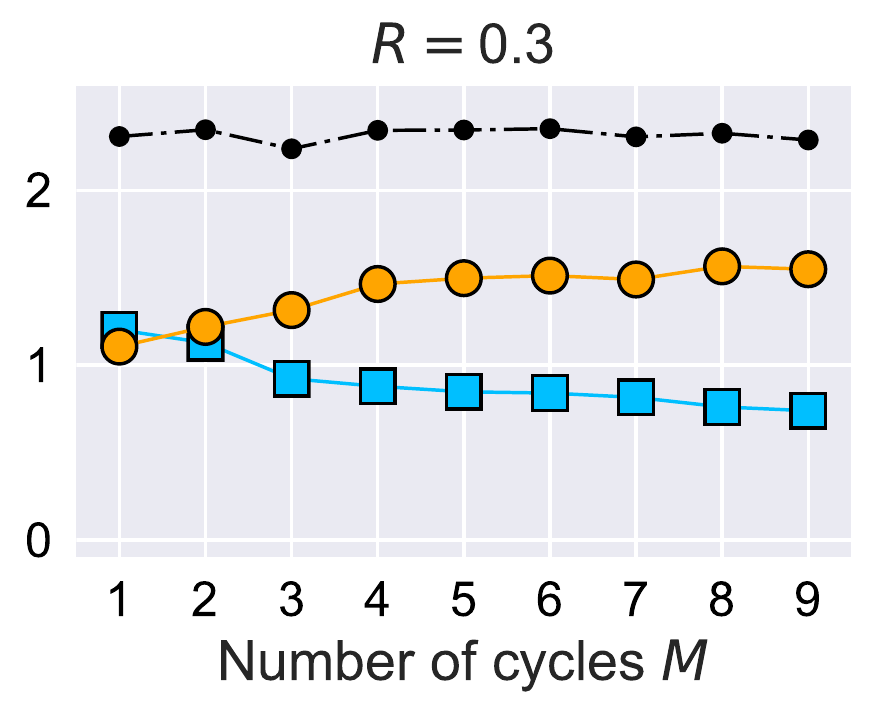} \\
		\hspace{-3mm}		
		\includegraphics[height=3.6cm]{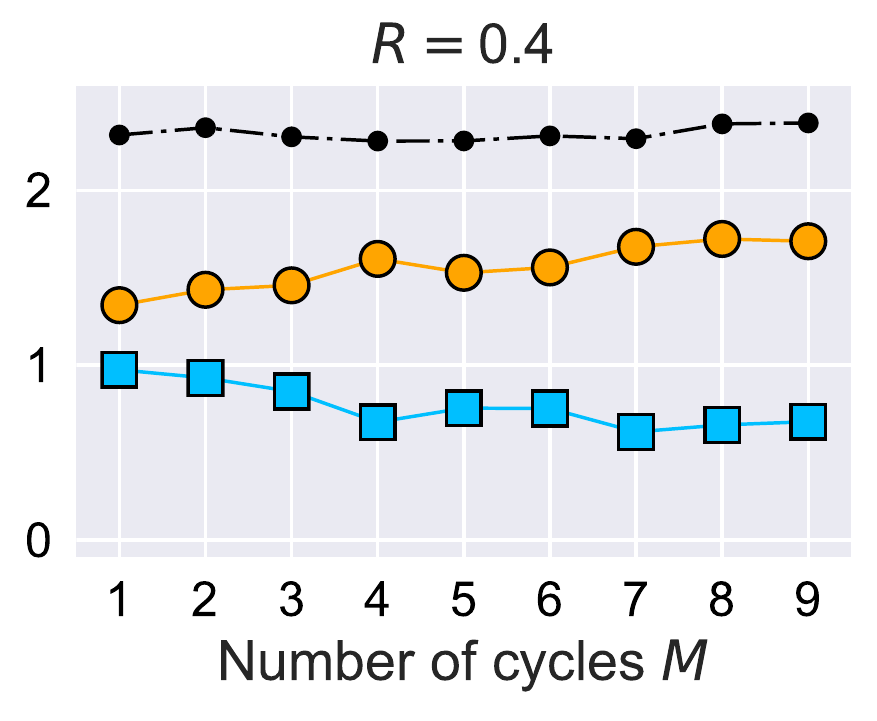} &
		\includegraphics[height=3.6cm]{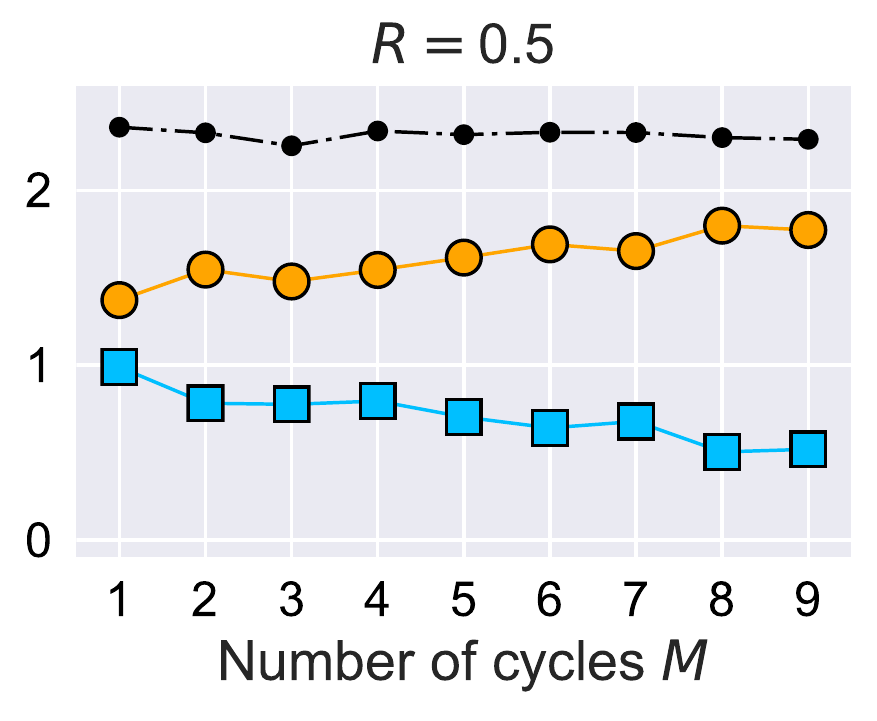} &
		\includegraphics[height=3.6cm]{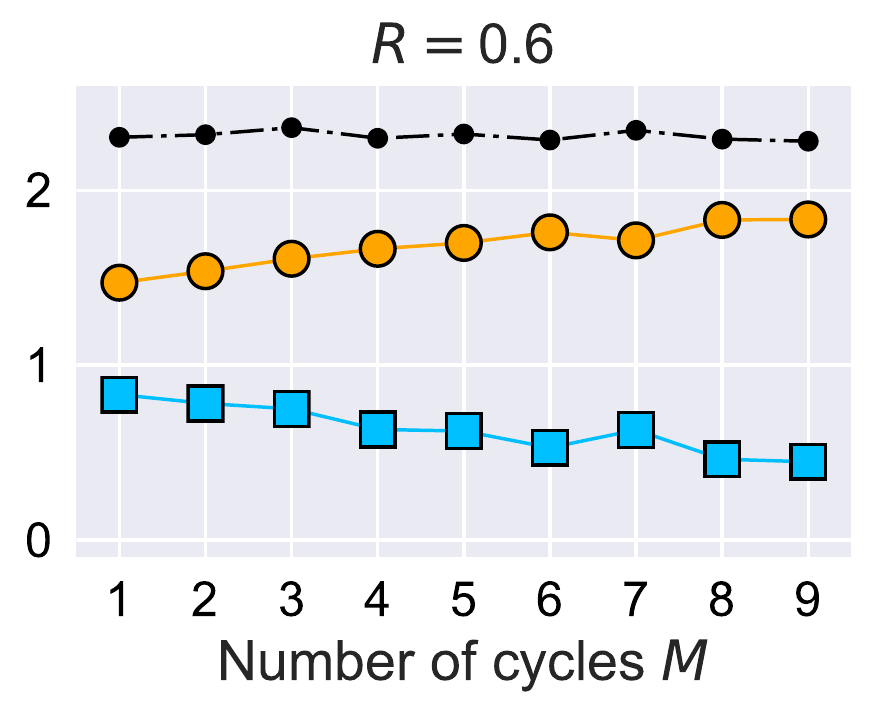} \\
		\hspace{-3mm}
		\includegraphics[height=3.6cm]{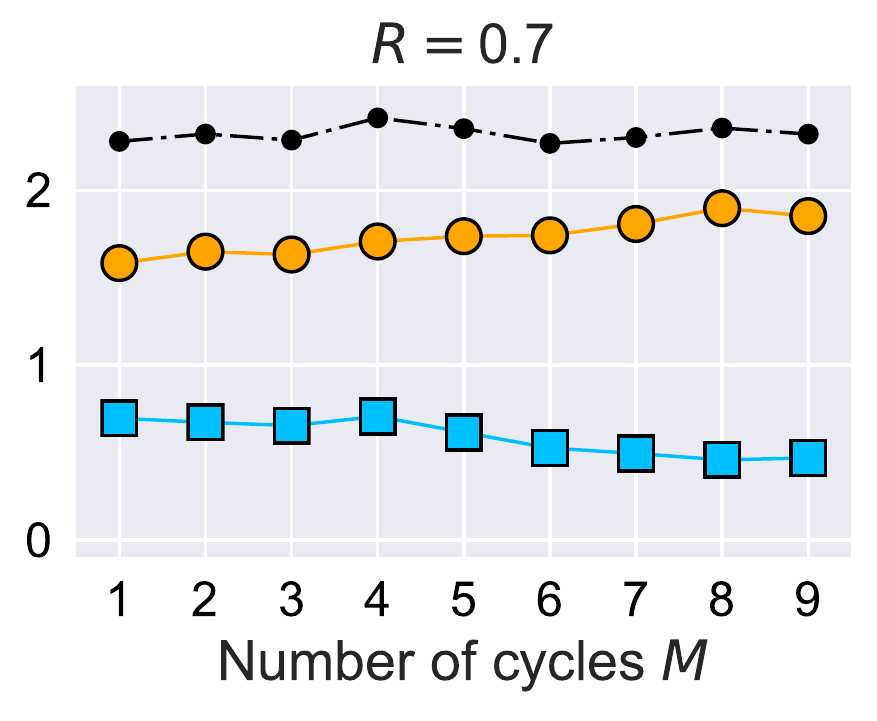} &
		\includegraphics[height=3.6cm]{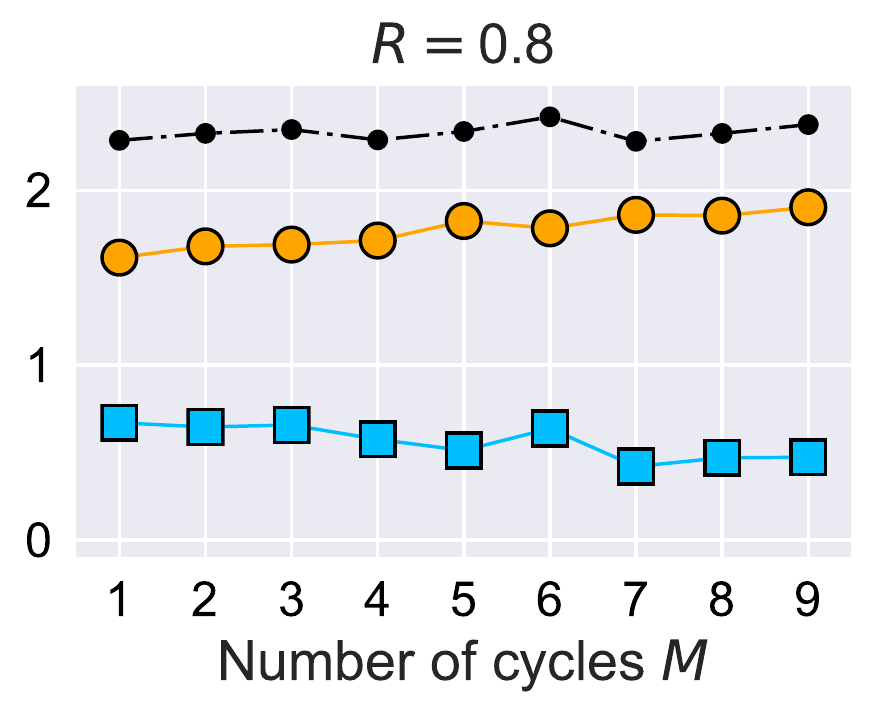} &
		\includegraphics[height=3.6cm]{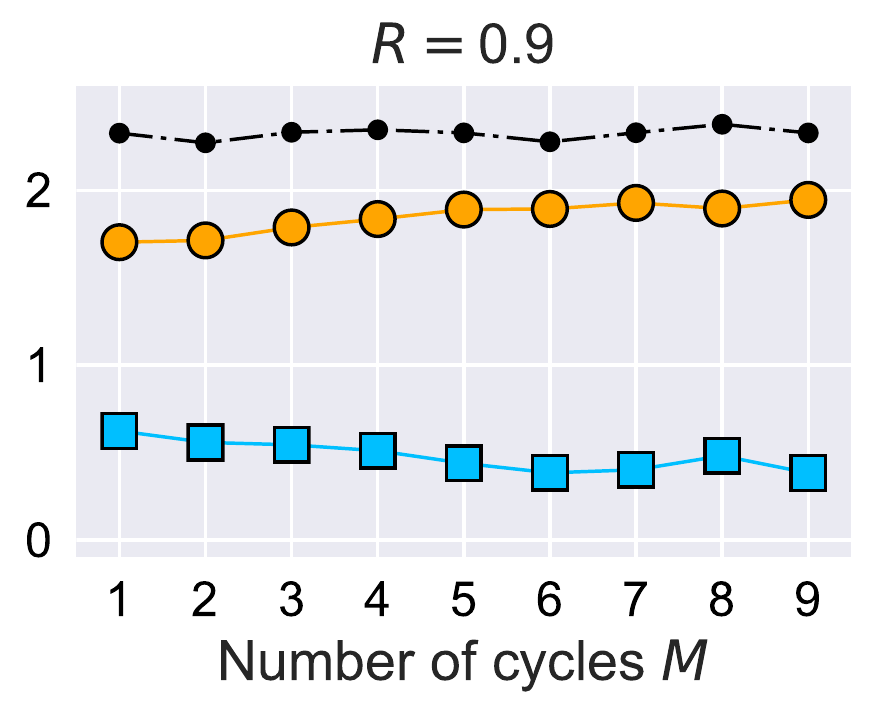} \\

		& \hspace{-20mm}
		\frame{
		\includegraphics[height=0.6cm]{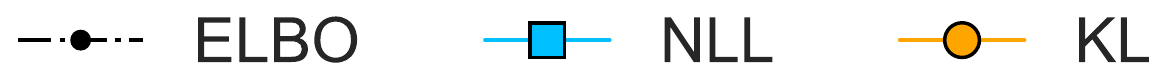}
		}
		 \hspace{-20mm}
		&
		% \multicolumn{3}{c}{}
	\end{tabular}
	\vspace{-2mm}
	\caption{The impact of hyper-parameter $M$: number of cycles. A larger number of cycles lead to better performance.
	The improvement is more significant when $R$ is small. The improvement is small when $R$ is large.}
	\vspace{-2mm}
	\label{fig:cycle_number_supp}
\end{figure*}

\begin{figure*}[t!]%\vspace{-25pt}
	\vspace{-0mm}\centering
	\begin{tabular}{ccc}
	    \hspace{-3mm}		
		\includegraphics[height=3.6cm]{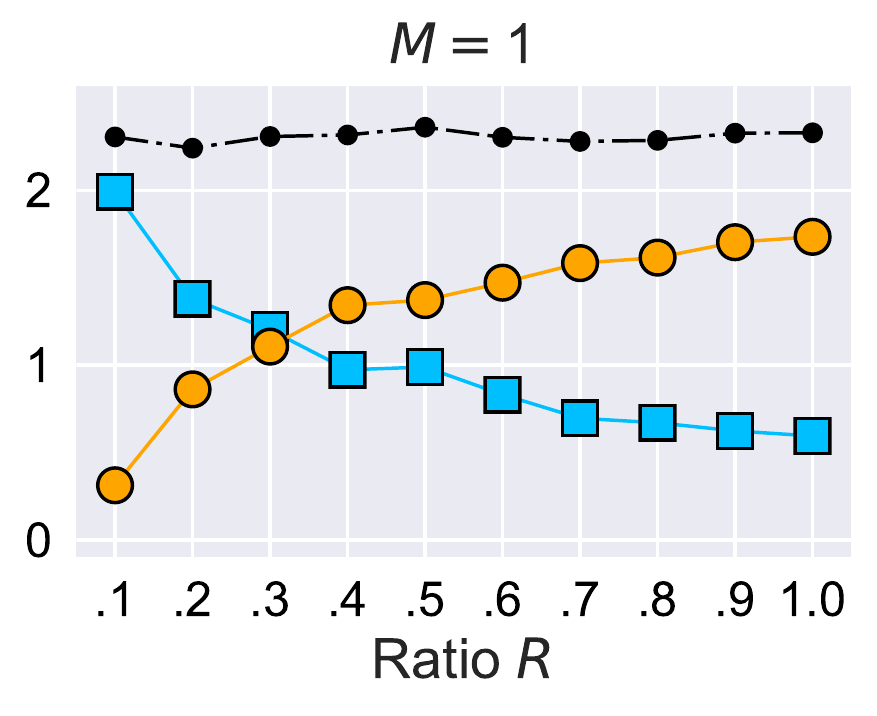} &
		\includegraphics[height=3.6cm]{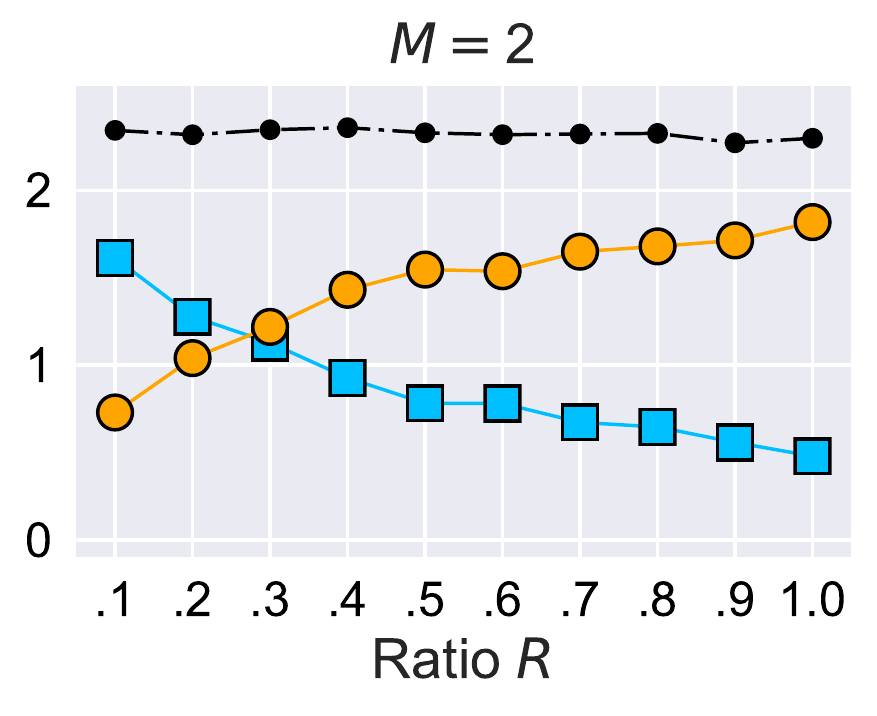} &
		\includegraphics[height=3.6cm]{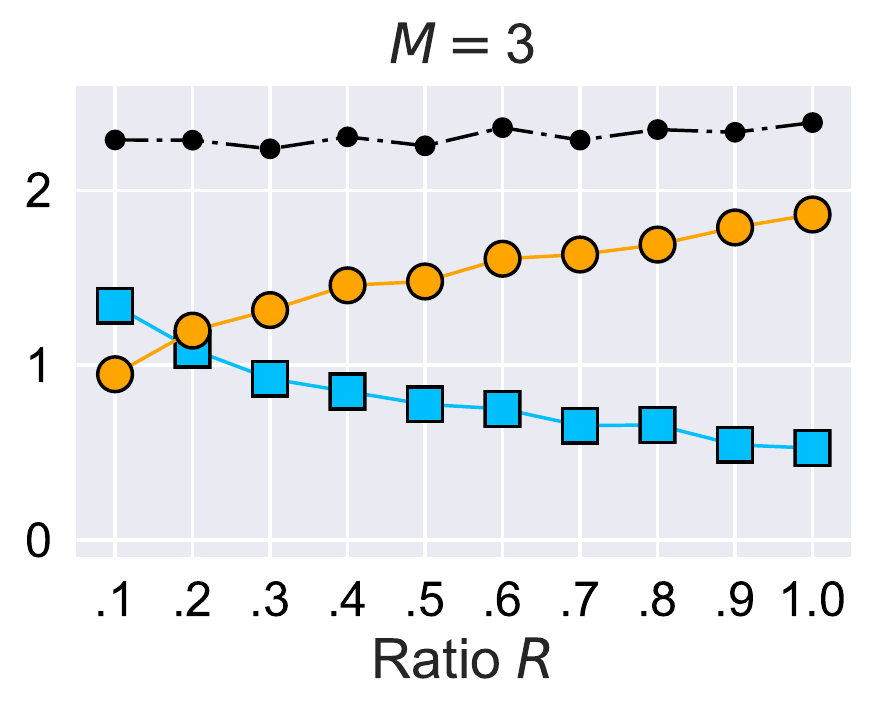} \\
		\hspace{-3mm}		
		\includegraphics[height=3.6cm]{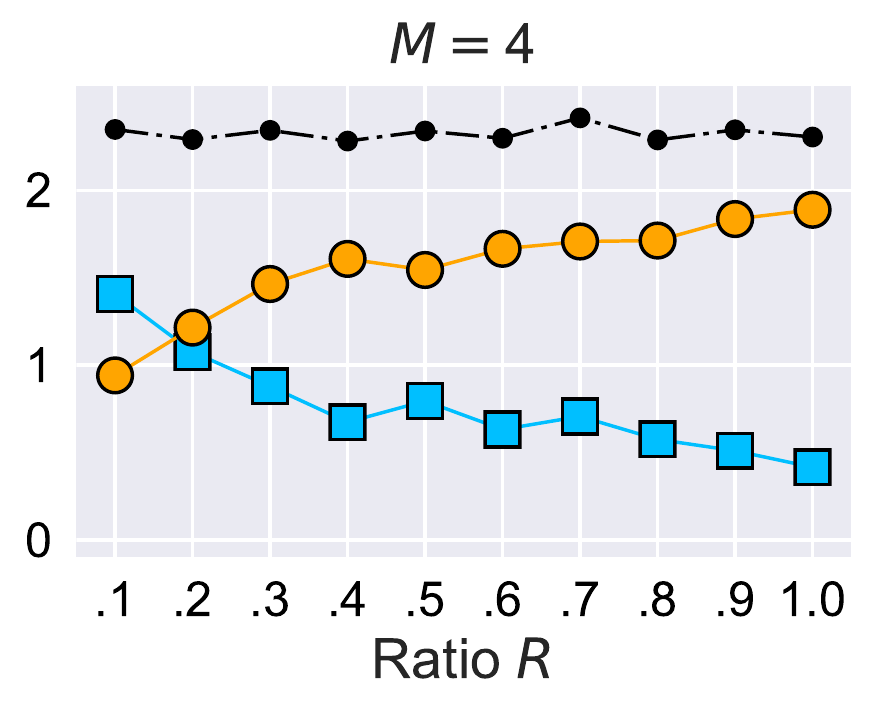} &
		\includegraphics[height=3.6cm]{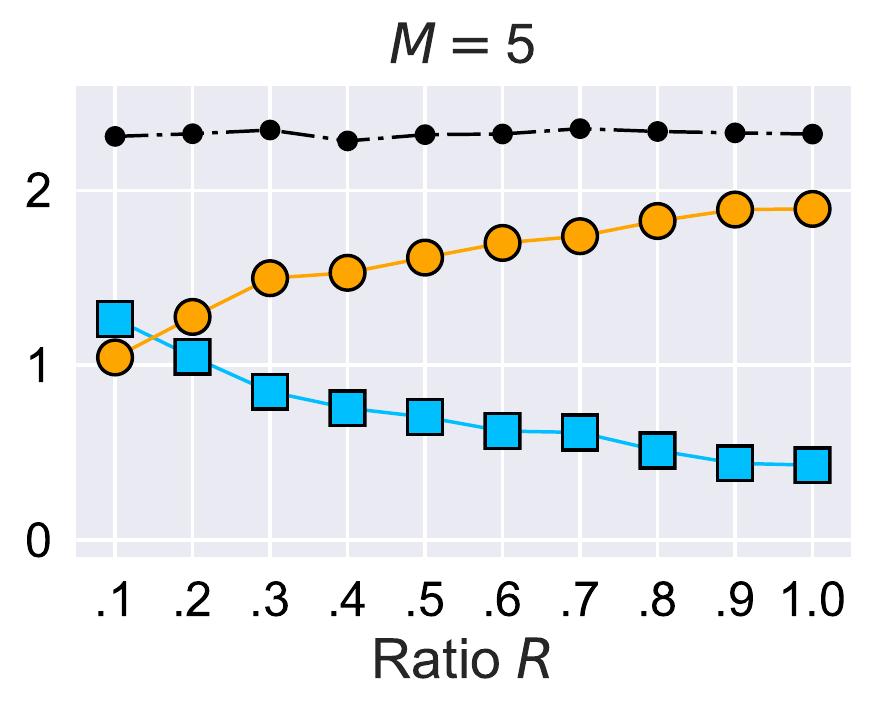} &
		\includegraphics[height=3.6cm]{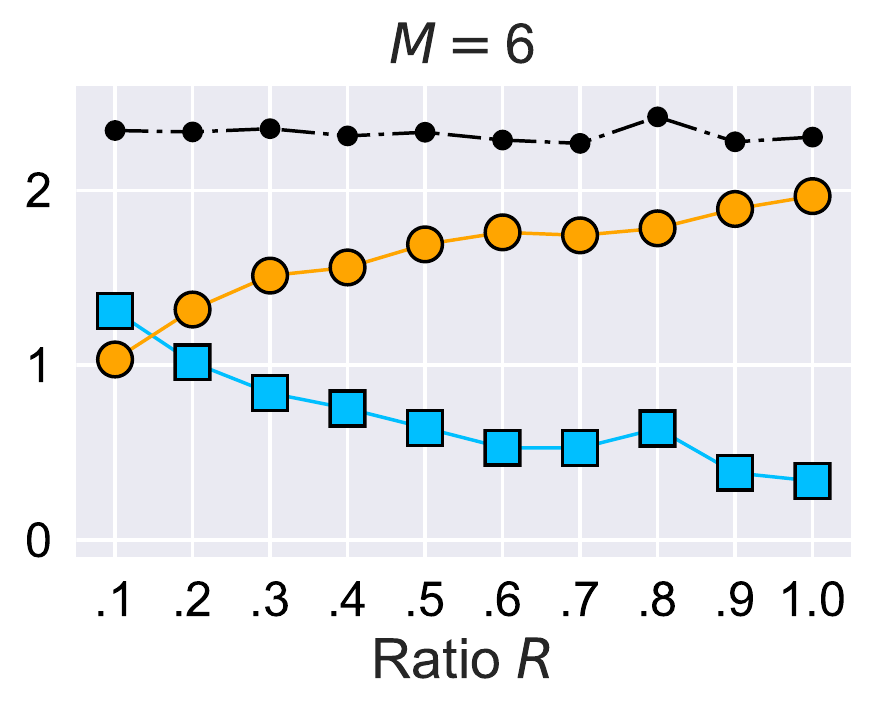} \\
		\hspace{-3mm}
		\includegraphics[height=3.6cm]{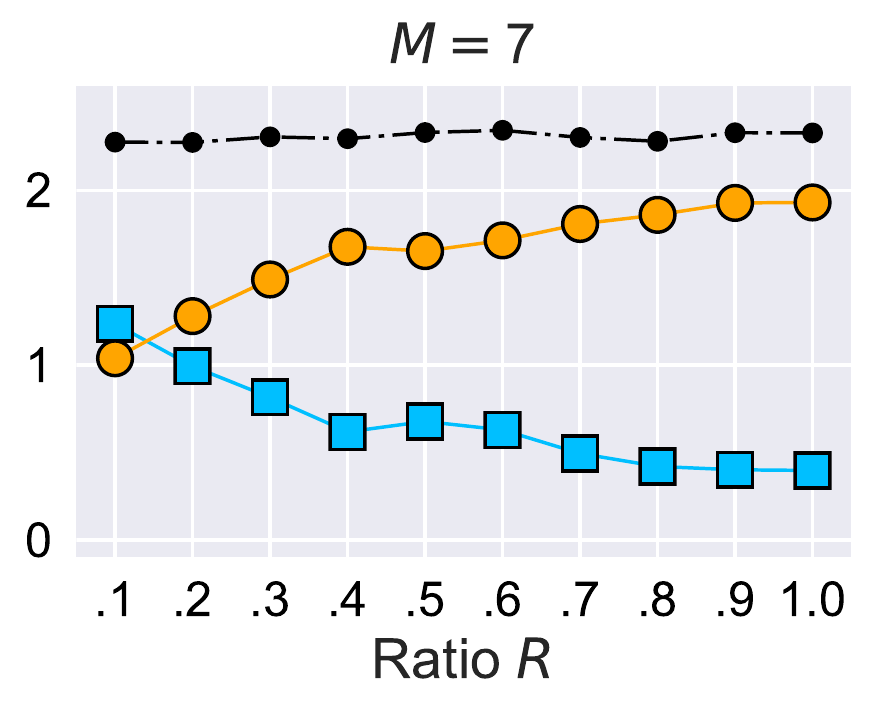} &
		\includegraphics[height=3.6cm]{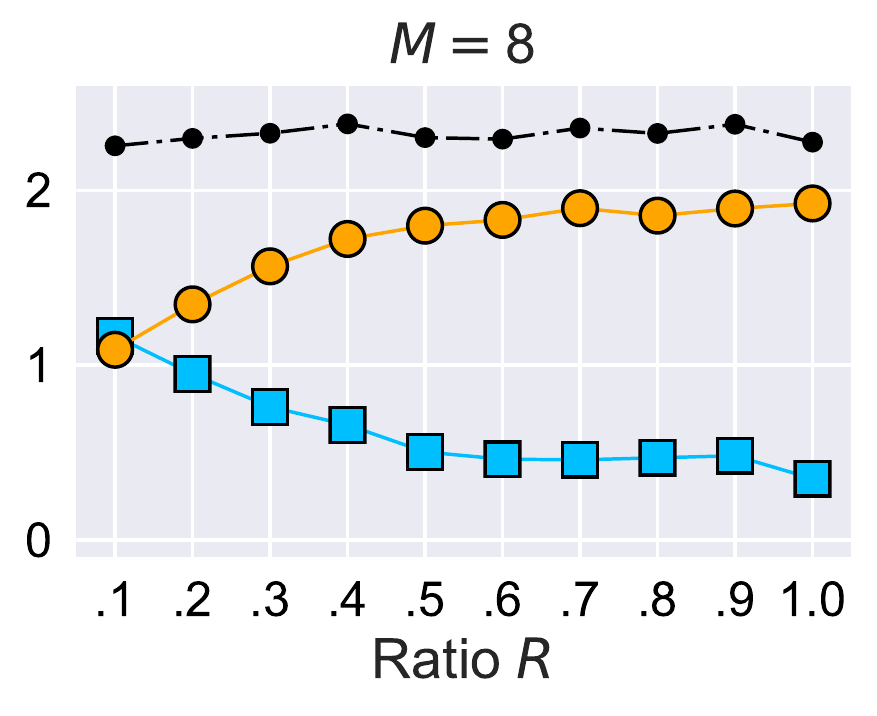} &
		\includegraphics[height=3.6cm]{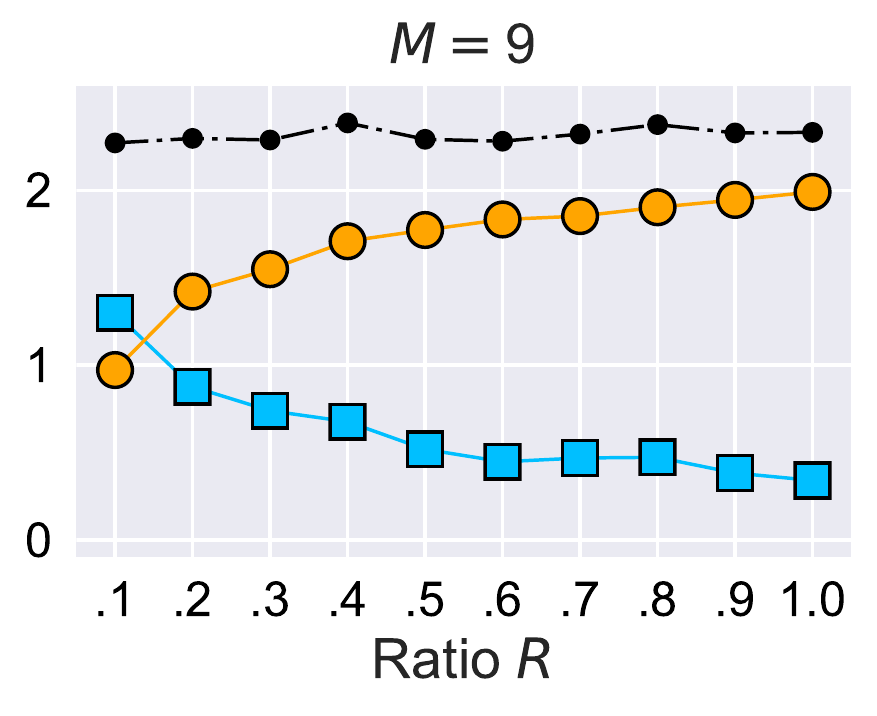} \\

		& \hspace{-20mm}
		\frame{
		\includegraphics[height=0.6cm]{figs/fig_hypara/legend.pdf}
		}
		 \hspace{-20mm}
		&
		% \multicolumn{3}{c}{}
	\end{tabular}
	\vspace{-2mm}
	\caption{The impact of hyper-parameter $R$: proportion for the annealing stage. A larger $R$ leads to better performance for various $M$. Small $R$ performs worse, because the schedule becomes more similar with constant schedule. $M=1$ recovers the monotonic schedule. Contrary to the convention that typically adopts small $R$, our results suggests that larger $R$ should be considered.}
	\vspace{-2mm}
	\label{fig:ratio_supp}
\end{figure*}

The cyclical schedule has two hyper-parameters $M$ and $R$. 
We provide the full results on $M$ and $R$ in Figure~\ref{fig:cycle_number_supp} and Figure~\ref{fig:ratio_supp}, respectively.
A larger number of cycles $M$ can provide higher performance for various proportion value $R$.

\begin{table}[t!]\centering
\begin{tabular}{l||c|c||c|c}
\hline
\textbf{Model} & \multicolumn{2}{c||}{CVAE} & 
\multicolumn{2}{c}{CVAE+BoW}  \\ \hline
\textbf{Schedule} & \textbf{M} & \textbf{C}
& \textbf{M} & \textbf{C} \\
\hline
B1 prec & 
0.326 & \textbf{0.423} & 0.384 & \textbf{0.397} \\
B1 recall &
0.214 & \textbf{0.391} & 0.376 & \textbf{0.387} \\
B2 prec & 
0.278 & \textbf{0.354} & 0.320 & \textbf{0.331} \\
B2 recall & 
0.180 & \textbf{0.327} & 0.312 & \textbf{0.323} \\
B3 prec & 
0.237 & \textbf{0.299} & 0.269 & \textbf{0.279} \\
B3 recall & 
0.153 & \textbf{0.278} & 0.265 & \textbf{0.275} \\
B4 prec & 
0.185 & \textbf{0.234} & 0.211 & \textbf{0.219} \\
B4 recall & 
0.122 & \textbf{0.220} & 0.210 & \textbf{0.219} \\
\hline
\end{tabular}
\caption{Comparison on dialog response generation. BLEU (B) scores 1-4 are used for evaluation. Monotonic (\textbf{M}) and Cyclical (\textbf{C}) schedules are tested on two models.}
\label{tab:dialog_supp}
\end{table}

\begin{figure*}[t!]%\vspace{-25pt}
	\vspace{-0mm}\centering
	\begin{tabular}{cc}
	    \hspace{-3mm}		
		\includegraphics[height=4.4cm]{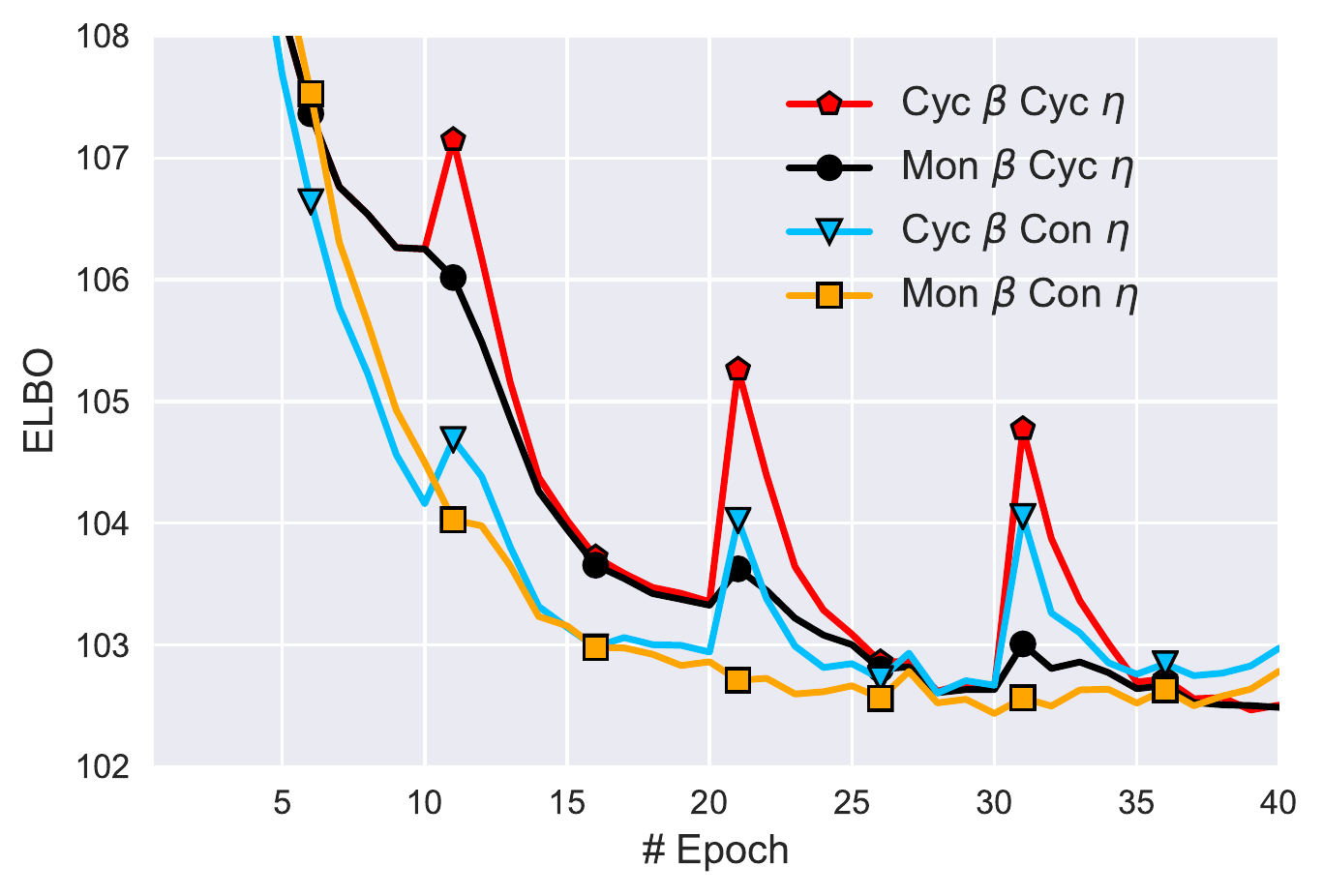} &
		\includegraphics[height=4.4cm]{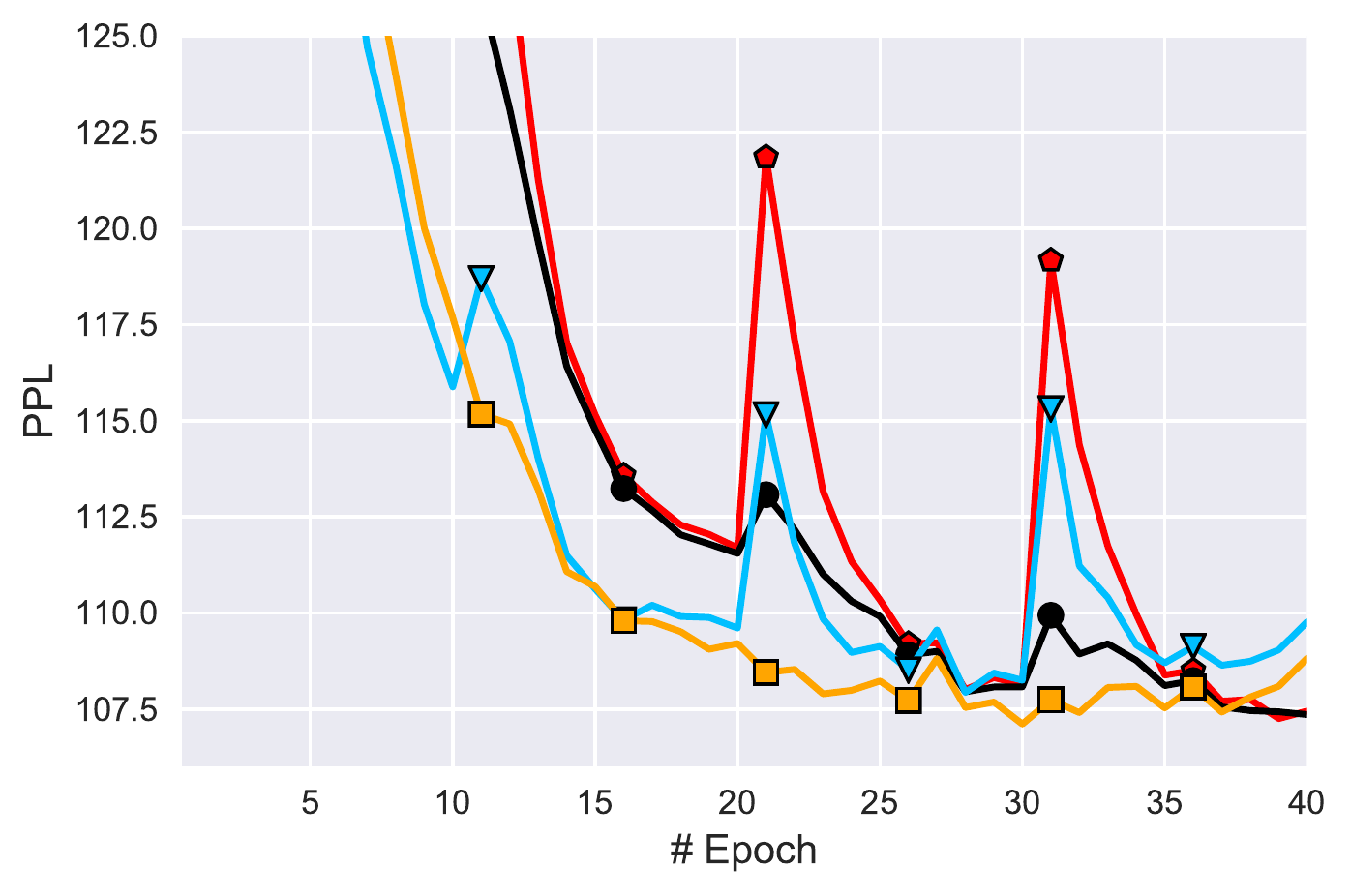} \\
		(a) ELBO &
		(b) PPL \\
		\includegraphics[height=4.4cm]{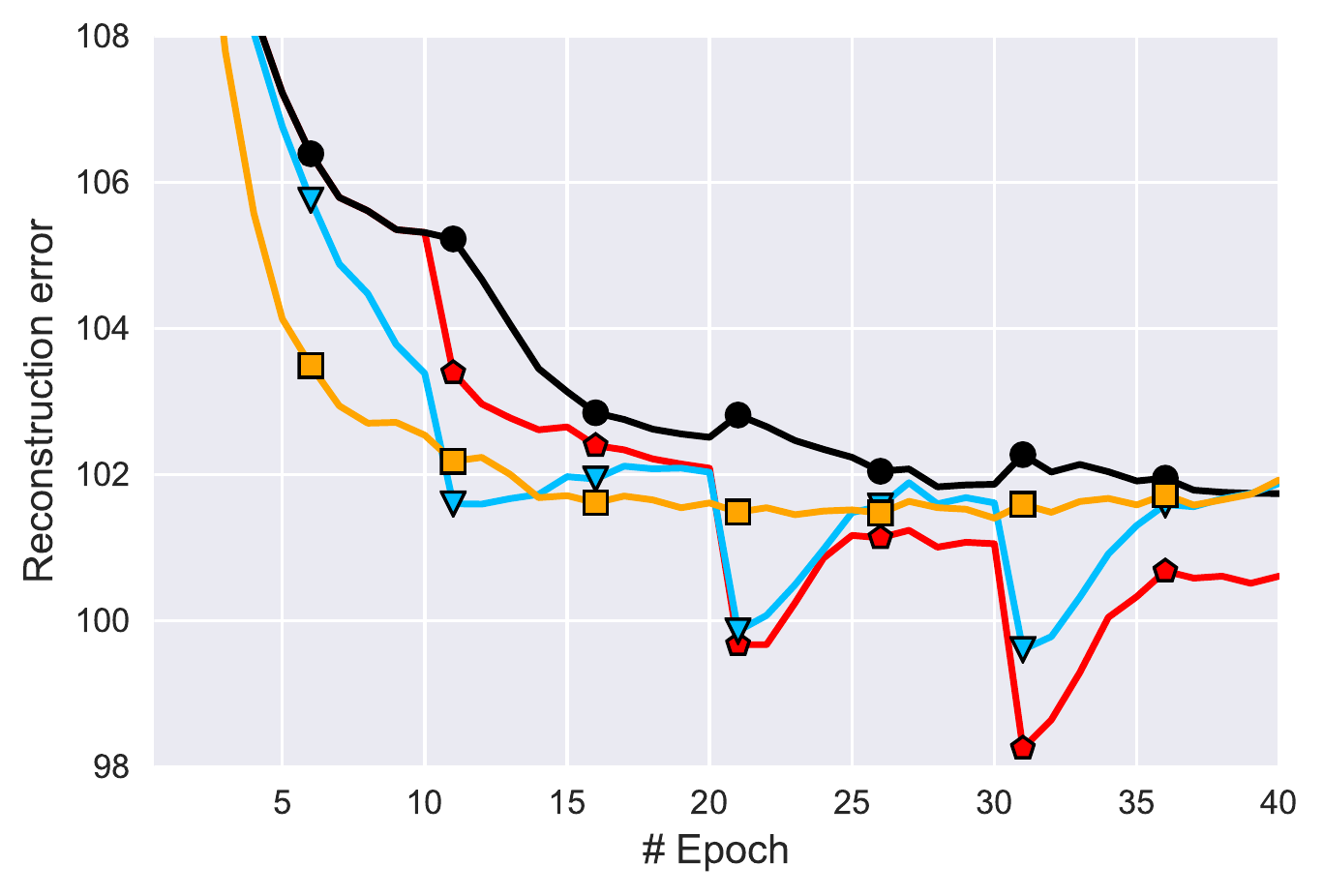} &		
		\includegraphics[height=4.4cm]{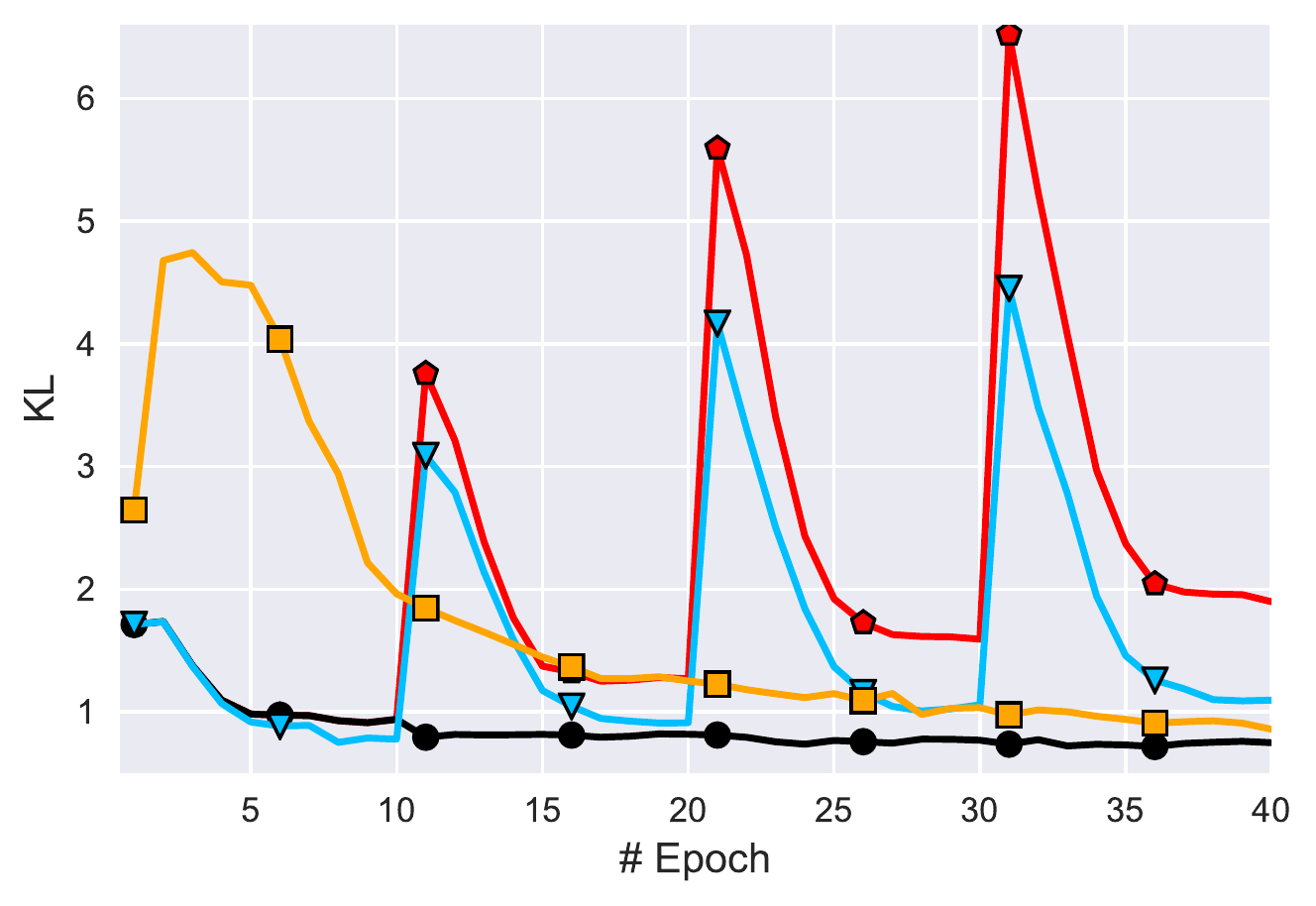} \\
		\hspace{-3mm}		
		(c) Reconstruction error &
		(d) KL
	\end{tabular}
	\vspace{-2mm}
	\caption{Ablation study on cyclical schedules on $\beta$ and $\eta$.}
	\vspace{-2mm}
	\label{fig:cyclical_eta_beta_supp}
\end{figure*}

% , because $\zv$ is not fully trained under VAE objective, and initialization used in warm restarts does not help much.

\end{document}